\documentclass[letterpaper]{article} 
\usepackage[preprint]{aaai2027}  
\usepackage[hyphens]{url}  
\usepackage{graphicx} 
\usepackage{natbib}  
\usepackage{caption} 
\usepackage{algorithm}
\usepackage{algpseudocode}

\usepackage{makecell}
\usepackage{multirow}
\usepackage{amsmath}
\usepackage{amssymb}

\usepackage{newfloat}
\usepackage{listings}
\DeclareCaptionStyle{ruled}{labelfont=normalfont,labelsep=colon,strut=off} 
\floatstyle{ruled}
\newfloat{listing}{tb}{lst}{}
\floatname{listing}{Listing}

\usepackage{booktabs}

\title{From Pixels to Hierarchical Sequences: Quadtree Mask Encoding\\for Vision-Language Binary Change Detection}
\author{
    Xiao An\textsuperscript{\rm 1}\equalcontrib,
    Ruikang Zhang\textsuperscript{\rm 2}\equalcontrib,
    Chen Zhong\textsuperscript{\rm 1},
    \\ Xuli Shen\textsuperscript{\rm 3},
    Jiaxing Sun\textsuperscript{\rm 3},
    Jiang Wu\textsuperscript{\rm 3}\corresponding,
    Wei He\textsuperscript{\rm 1}\corresponding
}
\affiliations{
    \textsuperscript{\rm 1}Wuhan University\quad
    \textsuperscript{\rm 2}Peking University\\
    \textsuperscript{\rm 3}Shanghai Artificial Intelligence Laboratory\\

}

\begin{document}

\maketitle

\begin{abstract}
Dense change detection in remote sensing requires vision-language models (VLMs) to compare bi-temporal images and generate accurate pixel-level masks. Existing VLMs are largely confined to change captioning outputs, and the few that produce pixel-level masks still rely on external decoders or flat text-as-mask serialization, which are less effective for small and fragmented changes. We introduce \textsc{QUAKE-CD}, a framework that recasts dense change prediction as syntax-verifiable structured generation. \textsc{QUAKE-CD} represents binary change masks as grammar-constrained quadtree token sequences, making the masks compact, syntactically checkable, and deterministically decodable within an autoregressive generation space. We further construct QUAKE-CoT, which pairs these sequences with chain-of-thought traces grounded in visual evidence, and jointly optimizes textual reasoning and spatial dense prediction through a progressive curriculum followed by grammar-gated dual-reward RL. On QUAKE-CoT, \textsc{QUAKE-CD} achieves $78.31\%$ accumulated F1, outperforming decoder-based and flat text-as-mask VLMs while producing more faithful bi-temporal reasoning.
\end{abstract}


\section{Introduction}
\label{sec:introduction}

Large vision-language models (VLMs) demonstrate profound capability in multimodal reasoning \cite{qwen3-vl,kimi-k2.5}, yet real-world applications increasingly demand spatially grounded outputs that delineate precise footprint boundaries \cite{vg-review,res-review}. Remote sensing change detection \cite{cd-review,changeclip} provides a stringent testbed for this imperative. Given bi-temporal images, models must synthesize cross-temporal visual evidence to render a pixel-level binary change map. Traditional change detection paradigms, including encoder--decoder networks and open-vocabulary formulations, narrowly optimize for discriminative mask prediction~\cite{changeclip,remotevar,semantic-cd}, structurally divorcing the semantic interpretation of changes from spatial footprints. While recent VLMs support interactive change captioning and localization \cite{changechat,btcchat,geollava}, \textit{natively} generating complete, dense spatial masks remains challenging. This limitation exposes a critical dichotomy between language-centric reasoning and pixel-level decision-making.

\begin{figure}[!t]
    \centering
    \includegraphics[width=0.94\linewidth]{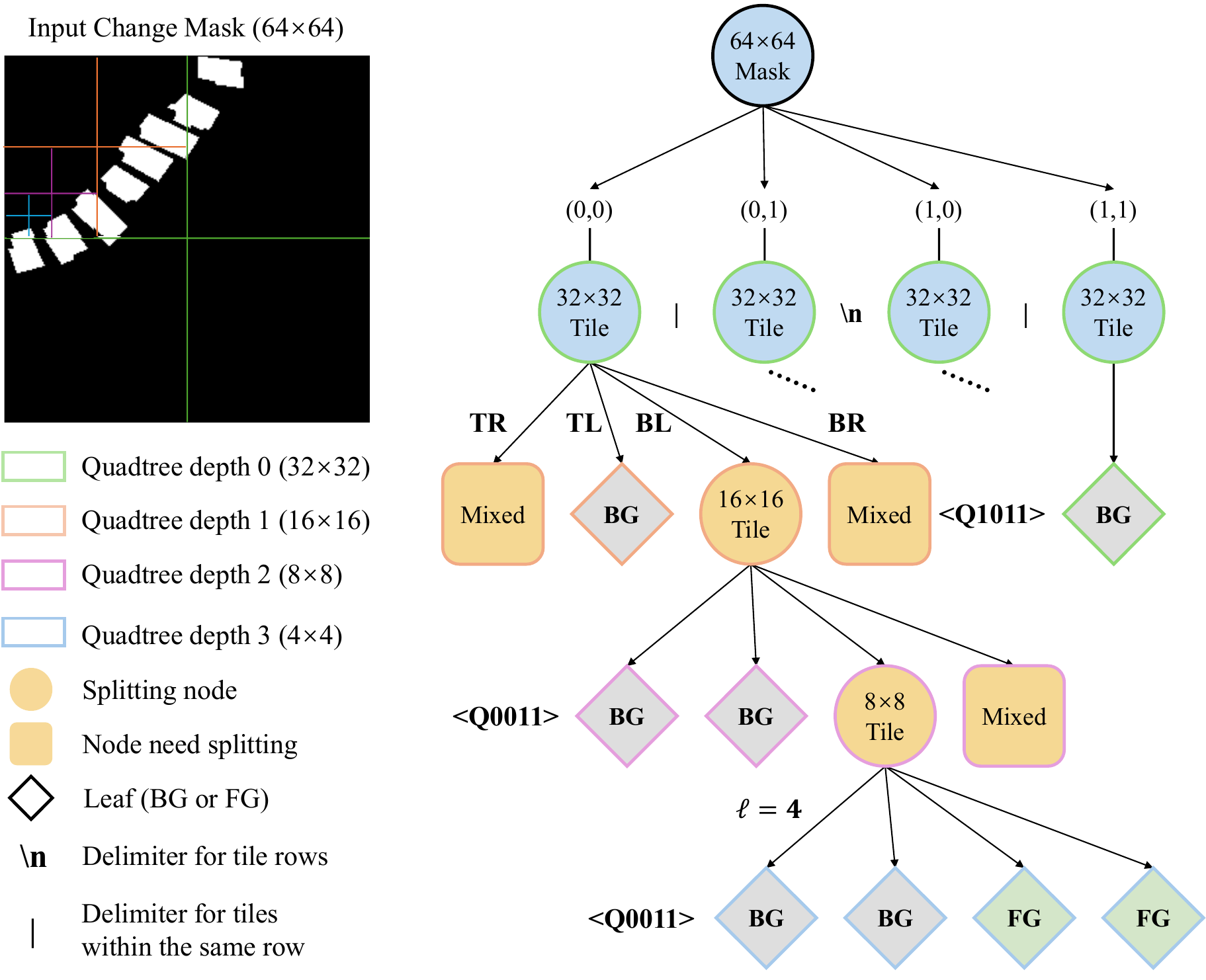}
    \caption{Illustration of the proposed \textit{QUAKE} grammar. We take $64\times 64$ change mask and $\ell=4$ as an example.}
    \label{fig:quake}
\end{figure}

Rather than bridging this gap, contemporary VLM architectures often relegate dense prediction to an auxiliary objective, mapping hidden states to spatial masks via task-specific decoders \cite{lisa,pixellm,vilacd,decode_the_delta}, which remains fundamentally orthogonal to the core autoregressive mechanism. While effective in isolation, this decoupling necessitates heterogeneous objectives, encumbering end-to-end training and stifling direct token-level optimization. Furthermore, sparse-cue methods that delegate mask generation to external segmenters (e.g., SAM~\cite{sam}) via bounding boxes or points~\cite{grasp,remotereasoner,rsthinker} not only lack native support for bi-temporal contexts but also falter on multiple fragmented instances. These coarse prompts inevitably entangle subtle targets with irrelevant background, triggering cascading localization errors that amplify false predictions.

Integrating dense prediction natively into VLMs requires formulating output masks as structured, autoregressive language. The emerging \textit{text-as-mask} paradigm \cite{text4seg,text4seg++,rsunivlm} offers a promising trajectory by serializing spatial layouts into textual descriptors. However, existing methods relying on flat patch-grid serialization treat all spatial regions equally, misaligning with sparse change masks where fine-grained boundaries demand high token density. Furthermore, this uniform encoding causes densely distributed targets in remote sensing (RS) imagery to suffer from severe spatial adhesion, erroneously collapsing distinct instances into a single connected component. Consequently, the task necessitates a hierarchical representation capable of efficiently compressing uniform regions while adaptively allocating tokens to areas demanding high spatial acuity.

We introduce \textsc{QUAKE-CD}, a framework that unifies language-centric reasoning with pixel-level prediction by encoding binary masks via \textbf{Qua}dtree mas\textbf{k} \textbf{E}ncoding (\textit{QUAKE})—a hierarchical spatial language shown in Figure~\ref{fig:quake} that supports dense spatial optimization directly within the native autoregressive token space. By recursively refining heterogeneous regions and collapsing uniform areas, \textit{QUAKE} adaptively allocates tokens to preserve fine-grained boundaries while compressing redundant background. Crucially, its grammar makes structural errors observable, transforming dense prediction into a syntax-verifiable language generation that aligns naturally with explicit reasoning traces.

Trained exclusively on \textit{QUAKE}, VLMs mechanically memorize the syntax but struggle to grasp the underlying semantics, degenerating into hollow pattern matching. To explicitly guide the model and anchor this grammar in genuine visual understanding, we construct \textit{QUAKE-CoT}, an instruction dataset that couples \textit{QUAKE} sequences with chain-of-thought (CoT) rationales detailing the type, location, and extent of changes. To avert optimization conflicts, our progressive curriculum decouples these dual objectives: the model first masters valid sequence generation before integrating the bi-temporal rationales. Exploiting the syntactic verifiability of \textit{QUAKE}'s textual form, we further align the model via grammar-gated dual-reward reinforcement learning (RL). A rule-based \textit{Quadtree Reward} validates syntax and scores decoded masks against ground truth using a Tversky-style spatial objective, preventing format collapse while optimizing dense prediction quality. In parallel, \textit{ThinkAnswer Reward}
evaluates whether the reasoning trace faithfully reflects the observed changes. Optimizing these rewards with GRPO \cite{grpo} jointly enforces grammaticality, pixel-level accuracy, and visually grounded reasoning.

Experiments show that \textsc{QUAKE-CD} substantially improves pixel-level VLM change detection. It achieves $78.31\%$ accumulated F1 and $73.18\%$ per-image F1, surpassing decoder-based baselines and uniform text-as-mask serializations. The results expose two failure modes of existing designs: decoder-based VLMs produce spatial outputs weakly coupled to cross-temporal evidence, while uniform grid serialization over-covers small, densely distributed changes. Reasoning evaluations further confirm that grounded spatial supervision yields more faithful bi-temporal explanations.

In summary, our main contributions are:
\begin{itemize}
    \item We introduce \textit{QUAKE}, a hierarchical text-as-mask language whose context-free grammar admits linear-time syntactic and structural validation, embedding dense spatial prediction into the autoregressive token space.
    \item We construct \textsc{QUAKE-CoT}, a dataset that explicitly couples hierarchical mask sequences with CoT rationales, and establish a progressive curriculum to anchor spatial grammar in genuine bi-temporal understanding.
    \item We develop a grammar-gated dual-reward RL paradigm that jointly optimizes format validity, pixel-level mask accuracy, and semantic reasoning fidelity within a unified token-level optimization process.
\end{itemize}

\section{Related Work}
\label{sec:related_works}

\noindent\textbf{Pixel-Level Prediction in RS Vision-Language Models.}
While VLMs have been widely adapted for RS applications including change captioning \cite{changechat}, change grounding \cite{changevg}, and RL-driven spatial reasoning \cite{remotereasoner, grasp}, achieving precise pixel-level interaction remains a systemic challenge. Most approaches either employ auxiliary mask decoders \cite{lisa,pixellm,segearth-r1} or generate sparse coordinate cues that inevitably delegate dense mask prediction to external segmenters \cite{grasp,rsthinker}. Although the emerging \textit{text-as-mask} paradigm bypasses external modules by serializing spatial layouts directly into text \cite{text4seg,rsunivlm}, its rigid, flat-grid formulation fails to capture fine boundaries and consistently induces severe spatial adhesion among dense semantic targets. \textsc{QUAKE-CD} fundamentally diverges by formulating the bi-temporal change mask as a grammar-constrained hierarchical \textit{QUAKE} sequence, adaptively concentrating tokens along boundaries to ensure native, precise mask generation without external visual dependencies.

\noindent\textbf{Grammar-Aligned Spatial Reasoning.}
Standard CoT prompting inherently struggles with complex spatial comprehension \cite{Spatial-MM}, routinely hallucinating rationales without explicit grounding supervision \cite{cure}. While recent advances employ grammar constraints for structural validity \cite{grammar-aligned-decoding} and RL to refine reasoning trajectories \cite{grpo}, these mechanisms operate in isolation: reasoning is evaluated as unstructured text, grammar superficially restricts decoding, and RL predominantly targets sparse bounding-box objectives \cite{grasp,remotereasoner}. \textsc{QUAKE-CD} departs from this line by inextricably anchoring the semantic rationale to a formal \textit{QUAKE} grammar; the resulting dual-reward RL framework jointly enforces grammatical determinism, semantic faithfulness, and pixel-level precision.
\section{Methodology}
\label{sec:methodology}

\subsection{Problem Formulation}
\label{sec:problem_formulation}

\textsc{QUAKE-CD} reformulates bi-temporal change detection as autoregressive structured generation. Given a bi-temporal image pair $(I_1, I_2)$, a VLM $\pi_\theta$ generates a tripartite token sequence $\mathbf{y} = [\mathbf{y}_{\mathrm{think}};\mathbf{y}_{\mathrm{answer}};\mathbf{y}_{\mathrm{seg}}]$, where $\mathbf{y}_{\mathrm{think}}$ constitutes a free-text CoT reasoning trace, $\mathbf{y}_{\mathrm{answer}}$ provides a concise change summary, and $\mathbf{y}_{\mathrm{seg}}$ acts as a quadtree encoding drawn from a formal language $\mathcal{L}_Q \subset V^*$. A deterministic decoder $\mathcal{D}: \mathcal{L}_Q \to \{0,1\}^{H \times W}$ then maps any valid encoding back to a binary change mask $\hat{M} = \mathcal{D}(\mathbf{y}_{\mathrm{seg}})$.

\begin{figure*}[!t]
    \centering
    \includegraphics[width=0.95\linewidth]{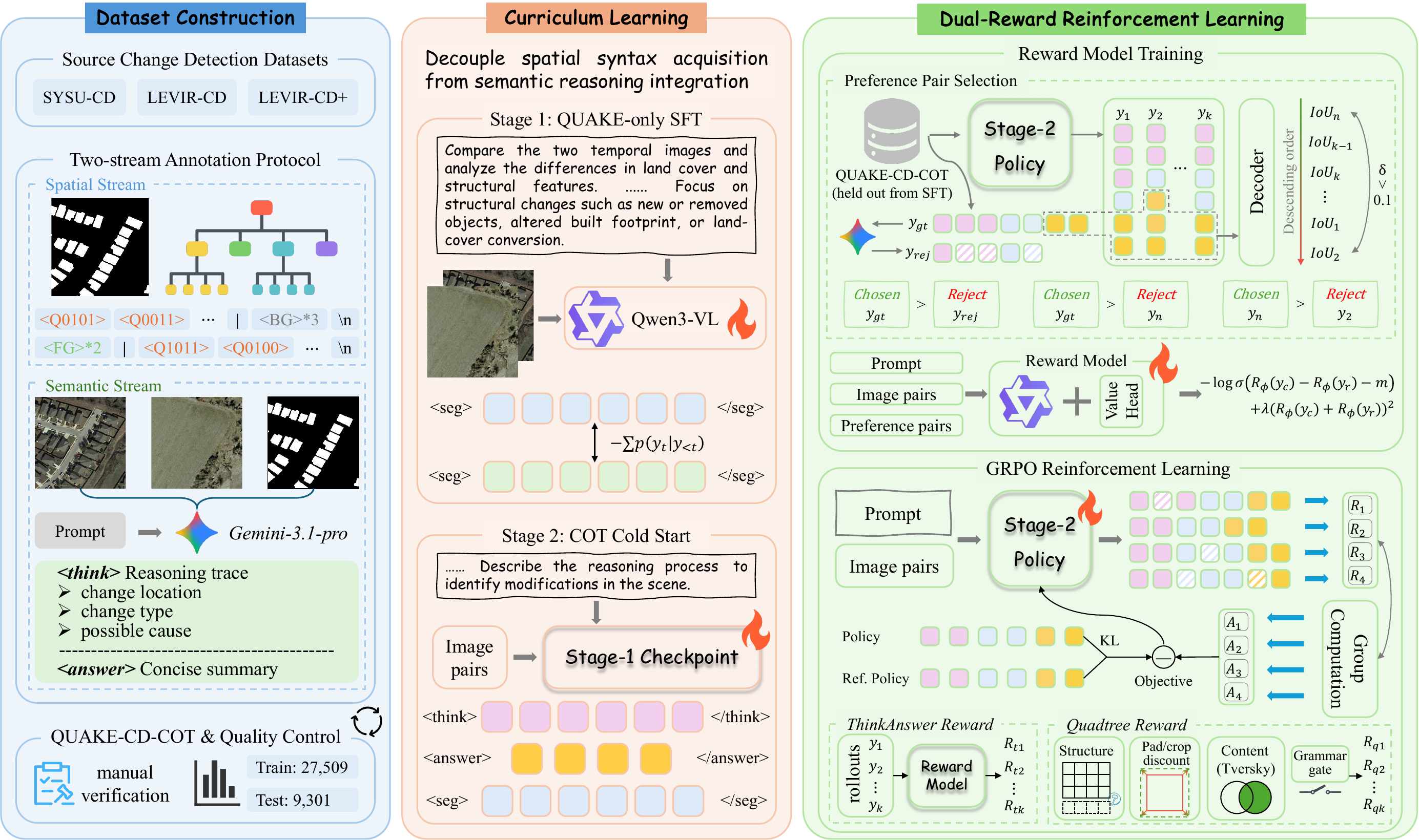}
    \caption{Illustration of the QUAKE-CD Pipeline. (1) Construction of the QUAKE-CoT dataset via a two-stream annotation protocol. (2) Two-stage progressive SFT curriculum. (3) Reward model training and grammar-gated dual-reward RL.}
    \label{fig:pipeline}
\end{figure*}

\subsection{Quadtree Mask Encoding}
\label{sec:quadtree_encoding}



\paragraph{Vocabulary and Encoding Grammar.}
The language $\mathcal{L}_Q$ is defined over an alphabet of special tokens integrated into the VLM vocabulary. This alphabet encompasses wrapper delimiters \texttt{<seg>} and \texttt{</seg>}, uniform tile tokens \texttt{<BG>} representing pure background and \texttt{<FG>} denoting pure foreground, alongside fifteen quadtree node tokens \texttt{<Q0001>} through \texttt{<Q1111>}. The all-zero pattern \texttt{<Q0000>} is explicitly excluded as it inherently signifies an empty quadrant.

The syntax of $\mathcal{L}_Q$ obeys a context-free grammar $\mathcal{G}_Q$. A valid encoding conforms to the structure \texttt{<seg>}\;G\;\texttt{</seg>}, where $G$ specifies a grid of non-overlapping $32{\times}32$ image tiles ordered via raster scan. Tiles within a row are delineated by \texttt{|} symbols, while distinct rows are separated by newline characters. This $32{\times}32$ tile dimension aligns perfectly with the native visual token granularity of Qwen3-VL \cite{qwen3-vl}, ensuring the output grid shares an identical spatial architecture with the visual feature map. Within each row, consecutive identical uniform tiles undergo run-length compression as \texttt{<BG>*$n$} or \texttt{<FG>*$n$} for integer $n \ge 2$.

As outlined in Figure~\ref{fig:quake} and Algorithm~\ref{alg:encode}, given a binary change mask $M \in \{0,1\}^{H \times W}$, a deterministic encoder $E: \{0,1\}^{H \times W} \to \mathcal{L}_Q$ partitions $M$ into the aforementioned tile grid. Each tile acts either as a uniform token or a quadtree expansion represented by a node \texttt{<Q}$abcd$\texttt{>}. The four constituent bits specify which respective quadrants contain foreground pixels across the top-left, top-right, bottom-left, and bottom-right sectors. Active quadrants undergo recursive preorder expansion until reaching a minimum leaf size $\ell$.

The grammar $\mathcal{G}_Q$ is context-free with bounded recursion depth $d_{\max} = \lceil\log_2(32/\ell)\rceil$. We further distinguish three levels of correctness: \textbf{syntactic validity}, certified in linear time by a single-pass parser under $\mathcal{G}_Q$; \textbf{structural validity}, verified during decoding by matching the reconstructed grid against the expected $H/32 \times W/32$ layout, and \textbf{semantic correctness}, optimized against the ground truth through the reward framework in Section~\ref{sec:dual_reward_rl}.
Crucially, this encoding is \textbf{recall-preserving} by construction: any leaf cell containing a foreground pixel dictates a foreground encoding, strictly guaranteeing $\forall (i,j)$ where $M[i,j]{=}1$, $\mathcal{D}(E(M))[i,j] = 1$. The complementary precision loss is strictly bounded by $\ell$, restricting over-encoding solely to boundary cells mixing foreground and background. The parameter $\ell$ offers explicit modulation of this precision--compression trade-off, with concrete value selections analyzed in Section~\ref{sec:experiments}.

\paragraph{Mask Decoding.}
The decoding function $\mathcal{D}: \mathcal{L}_Q \to \{0,1\}^{H \times W}$ effectively reverses $E$ by parsing the \textit{QUAKE} sequence, resolving run-length compression, and recursively rendering quadtree nodes back into $32{\times}32$ pixel blocks. Consequently, $\mathcal{D}$ operates as a purely \textbf{non-parametric} mapping, circumventing neural networks and ensuring high efficiency during both RL reward computation and evaluation.

\subsection{Supervised Fine-Tuning with Curriculum}
\label{sec:sft_curriculum}


\paragraph{Dataset Construction.}
Existing change detection benchmarks, SYSU-CD~\cite{sysu-cd}, LEVIR-CD and LEVIR-CD+~\cite{levir-cd}, provide pixel-level masks but no language annotations. We assemble QUAKE-CoT from these sources via a two-stream annotation protocol. As illustrated in Figure~\ref{fig:pipeline}, the \textbf{spatial stream} deterministically encodes each ground-truth mask as $E(M) \in \mathcal{L}_Q$. The \textbf{semantic stream} prompts \textit{Gemini-3.1-Pro-Thinking-Preview}~\cite{gemini}, conditioned on the bi-temporal pair and ground-truth mask, to produce a reasoning trace covering change type, location, extent, and plausible cause, followed by a concise summary. Dataset details and system prompts are shown in Appendix~\ref{sec:appendix_dataset} and~\ref{sec:appendix_prompt}.

These independently generated streams are subsequently paired into samples of the form \texttt{<think>} \textit{reasoning} \texttt{</think>} \texttt{<answer>} \textit{summary} \texttt{</answer>} \texttt{<seg>} \textit{QUAKE} \texttt{</seg>}. Mask-conditioned generation makes these traces a \textbf{grounding supervision signal} that anchors \textit{QUAKE} syntax in visually verifiable semantics, not a record of naturalistic inference. The original training splits yield 27{,}509 training samples; the union of validation and test splits forms a 9{,}301-sample test set, precluding leakage.
Quantitative assessments regarding the round-trip fidelity of $E$ and $\mathcal{D}$ on the QUAKE-CoT dataset are established in Table~\ref{tab:appendix_encoding_fidelity}.
To our knowledge, QUAKE-CoT constitutes the first instruction-tuning dataset for bi-temporal change detection that pairs structured reasoning traces with formalized, deterministically decodable dense mask encodings.

\noindent\textbf{Curriculum Stage-1: QUAKE-Only SFT.}
Attempting simultaneous acquisition of \textit{QUAKE} grammar alongside free-text reasoning precipitates destructive interference, forcing the model to fracture capacity between syntax mastery and semantic analysis. \textit{Stage-1} intentionally isolates spatial syntax acquisition. We fine-tune the model to exclusively generate the \texttt{<seg>} block conditioned on a bi-temporal image pair, dedicating its entire generation budget to mastering grid dimensions, uniform tile semantics, recursive node expansion, run-length encoding, and row delimiters.

\noindent\textbf{Curriculum Stage-2: CoT Cold-Start.}
\textit{Stage-2} integrates reasoning by fine-tuning the Stage-1 checkpoint to synthesize the complete tripartite output. This \textit{cold-start} strategy successfully preserves the entrenched \textit{QUAKE} syntax while catalyzing the acquisition of cross-temporal reasoning capabilities. Subsequent ablations in Section~\ref{sec:exp_ablation} validate the efficacy of this curriculum. The sequential dependency linking these stages operationalizes the curriculum's central principle: demanding mastery of the spatial alphabet before permitting the composition of spatially grounded visual explanations.

\subsection{Grammar-Gated Dual-Reward RL}
\label{sec:dual_reward_rl}

While SFT yields syntactically valid \textit{QUAKE} sequences paired with grounded reasoning, it fundamentally lacks the mechanism to directly optimize the composite objective encompassing format validity, pixel-level accuracy, and reasoning faithfulness. We resolve this limitation with a dual-reward RL framework to jointly optimize dense spatial outputs and strengthen the reasoning trace as a semantic guide for \textit{QUAKE} generation within the autoregressive space.

\noindent\textbf{Reward Model Construction.}
To train a dedicated reward model for evaluating semantic reasoning faithfulness, we construct preference pairs from the QUAKE-CoT split held out from SFT. The cold-start model generates multiple completions per sample under high-temperature decoding. Each completion's corresponding \textit{QUAKE} sequence is decoded via $\mathcal{D}$ and scored against the ground-truth mask $M$ using Intersection over Union (IoU). In parallel, we synthesize one hard-negative response $y_{\mathrm{neg}}$ per sample by prompting \textit{Gemini-3.1-Pro-Thinking-Preview} to perturb core semantic facts while preserving fluency and the grammatical structure.

Let $s(y)=\mathrm{IoU}(\mathcal{D}(y_{\mathrm{seg}}), M)$ denote the decoded IoU of a rollout. Given a candidate pool $\mathcal{C} = \{y_{\mathrm{gt}}\} \cup \{y_k\}_{k=1}^{K} \cup \{y_{\mathrm{neg}}\}$, where $y_{\mathrm{gt}}$ is the ground-truth reasoning trace, and $\{y_k\}_{k=1}^{K}$ denotes $K$ empirical rollouts, we form preference pairs following the five-tier hierarchy in Appendix~\ref{sec:appendix_rm}, with $y_{\mathrm{gt}}$ and $y_{\mathrm{neg}}$ serving as fixed top and bottom anchors, and retain rollout pairs whose IoU gap exceeds the threshold $\delta$:
\begin{equation}
\mathcal{D}_{\mathrm{pref}} = \{(y_c, y_r) \in \mathcal{C} \times \mathcal{C} \mid y_c \succ y_r\},
\label{eq:pref}
\end{equation}
The reward model $R_\phi$ is trained on $\mathcal{D}_{\mathrm{pref}}$ with Qwen3-VL-8B-Instruct under the objective:
\begin{equation}
  \begin{aligned}
  \mathcal{L}_{\mathrm{RM}}
  = -\mathbb{E}_{\mathcal{D}_{\mathrm{pref}}}\big[
  &\log \sigma(R_\phi(y_c) - R_\phi(y_r)) \\
  &- \lambda (R_\phi(y_c) + R_\phi(y_r))^2
  \big],
  \end{aligned}
  \label{eq:rm}
\end{equation}
where $y_c$ and $y_r$ indicate chosen and rejected responses, respectively, and $R_\phi$ evaluates the semantic caliber of the \texttt{<think>}\texttt{<answer>} segments. Details on negative generation and RM training dynamics are provided in \textit{\textbf{Appendix~\ref{sec:appendix_rm}}}.

\begin{table*}[!t]
\centering
\setlength{\aboverulesep}{0pt}
\setlength{\belowrulesep}{0pt}
\begin{tabular}{lccccccc}
\Xhline{1.2pt}
\multirow{2}{*}{Method} & \multicolumn{3}{c}{Accumulated Mean}             & \multicolumn{3}{c}{Per-Image Mean}               & Throughput \\ \cmidrule(lr){2-4}\cmidrule(lr){5-7}
                & Precision & Recall & F1    & Precision & Recall & F1    & (pairs/s) \\ \Xhline{1.2pt}
\textcolor{gray}{BIT}          & \textcolor{gray}{77.08} & \textcolor{gray}{76.80} & \textcolor{gray}{76.94} & \textcolor{gray}{66.34} & \textcolor{gray}{66.64} & \textcolor{gray}{63.19} & \textcolor{gray}{50.25} \\
\textcolor{gray}{ChangeFormer} & \textcolor{gray}{84.23} & \textcolor{gray}{78.84} & \textcolor{gray}{81.44} & \textcolor{gray}{76.61} & \textcolor{gray}{71.06} & \textcolor{gray}{70.88} & \textcolor{gray}{65.78}  \\
\textcolor{gray}{ChangeCLIP}   & \textcolor{gray}{84.32} & \textcolor{gray}{78.50} & \textcolor{gray}{81.30} & \textcolor{gray}{57.31} & \textcolor{gray}{50.58} & \textcolor{gray}{51.04} & \textcolor{gray}{74.93} \\ \hline
LISA-7B         & 16.65     & 73.42  & 27.14 & 18.04     & 46.83  & 21.91 & 40.21     \\
LISA-llama2-13B & 15.04     & 76.45  & 25.13 & 17.10     & 49.20  & 21.45 & 27.15     \\
GLaMM           & 14.85     & 52.07  & 23.11 & 16.22     & 29.59  & 17.44 & 18.77     \\
PixelLM         & 16.79     & 72.51  & 27.27 & 17.39     & 46.38  & 21.38 & 14.01     \\ \hline
Text4Seg        & 14.52     & 84.07  & 24.76 & 15.27     & 54.88  & 19.65 & \textbf{86.54}     \\
RSUniVLM        & 50.12     & 82.45  & 62.34 & 48.76     & 81.23  & 60.94 & 82.30     \\ \hline
\textbf{QUAKE-CD}       & \textbf{70.99} & \textbf{87.31} & \textbf{78.31} & \textbf{68.77} & \textbf{84.09} & \textbf{73.18} & 74.27 \\ \Xhline{1.2pt}
\end{tabular}
\caption{Dense change detection performance on the QUAKE-CoT test set. We compare against specialist change detectors (BIT, ChangeFormer, and ChangeCLIP), decoder-based pixel-level VLMs (LISA, GLaMM, PixelLM), a flat text-as-mask method (Text4Seg), and a remote sensing VLM (RSUniVLM) with native bi-temporal input and text-as-mask serialization.}
\label{tab:change_detection}
\end{table*}

\noindent\textbf{ThinkAnswer Reward.}
To systematically enforce the semantic validity of the underlying logic during RL, the \textit{ThinkAnswer Reward} $R_t$ acts as a direct application of the previously trained reward model $R_\phi$. It specifically scores the \texttt{<think>} reasoning trace and \texttt{<answer>} summary segments. This incorporates query-wise normalization to guarantee stable optimization signals across highly diverse prompts. Operating as a crucial semantic regularizer, this reward signal discourages structurally valid but factually unfaithful explanations from receiving inflated high scores.

\noindent\textbf{Quadtree Reward.}
The \textit{Quadtree Reward} $R_q$ operates as a rigorous rule-based function evaluating spatial mask quality natively from the generated sequence. A \textit{grammar gate} validates the \texttt{<seg>} sequence against $\mathcal{G}_Q$ via a single-pass linear-time parser. Concurrently, a \textit{structure score} mandates grid dimension alignment with the ground truth. The subsequent \textit{content score} calculates the Tversky index~\cite{tversky} between the decoded prediction $\hat{M}$ and ground truth $M$.
Additionally, a \textit{pad/crop discount} penalizes topological dimension mismatches necessitating fallback alignment, while an \textit{empty-GT guard} outright neutralizes the trivial exploitation strategy of perpetually predicting pure background (see details in \textit{\textbf{Appendix~\ref{sec:appendix_dual_reward}}}).

\noindent\textbf{GRPO Optimization.}
We optimize the VLM policy starting from the cold-start checkpoint utilizing the GRPO objective~\cite{grpo}, using the reward:
\begin{equation}
R(\mathbf{y}, M) = (1-\mu) R_{\mathrm{q}}(\mathbf{y}, M) + \mu R_{\mathrm{t}}(I_1,I_2,\mathbf{y})
\label{eq:grpo}
\end{equation}
under KL regularization against the reference policy $\pi_{\mathrm{ref}}$, where $\mu$ dictates the weight of the semantic reward. This dual-reward architecture instantiates the central hypothesis of this work: formulating dense spatial outputs as structured language enables joint, token-level optimization of format validity, pixel-level accuracy, and reasoning faithfulness.

\section{Experiments}
\label{sec:experiments}

\subsection{Experimental Setup}
\label{sec:exp_setup}

\noindent\textbf{Setup.}
All experiments utilize the QUAKE-CoT established in Section~\ref{sec:sft_curriculum}. We allocate 80\% of the training set to the two-stage curriculum and reserve the remaining 20\% exclusively for reward model training and RL. The framework is initialized from Qwen3-VL-8B-Instruct~\cite{qwen3-vl}. Training is executed via ms-swift~\cite{ms-swift} on eight NVIDIA H200 GPUs, while inference is served through vLLM~\cite{vllm}. Both SFT stages and the reward model employ LoRA~\cite{lora} with rank~$64$, with the reward model initialized from a fresh backbone copy. The RL phase optimizes the policy via GRPO~\cite{grpo} using eight rollouts per prompt, and a KL coefficient of $0.04$. Appendix~\ref{sec:appendix_hparams} catalogs the comprehensive hyperparameter specifications.

\noindent\textbf{Baselines.}
We compare against three complementary baseline groups. First, we include representative specialist remote sensing change detectors: BIT~\cite{bit}, ChangeFormer~\cite{changeformer}, and ChangeCLIP~\cite{changeclip}. Second, we evaluate pixel-level VLMs across two prevailing paradigms. LISA~\cite{lisa}, GLaMM~\cite{glamm}, and PixelLM~\cite{pixellm} perform decoder-based mask prediction, while Text4Seg~\cite{text4seg} and RSUniVLM~\cite{rsunivlm} serialize masks autoregressively. Specialist change detectors and RSUniVLM are retrained on the QUAKE-CoT training split. RSUniVLM natively accepts RS bi-temporal pairs, while the remaining pixel-level VLMs assume single-image inputs. Following~\cite{calico,mc-bench}, we concatenate the bi-temporal pair $(I_1, I_2)$ along the width dimension and fine-tune them on the QUAKE-CoT training split under their original training recipes. Third, for reasoning-quality evaluation, we include Teochat~\cite{teochat} and EarthDial~\cite{earthdial}. These temporal RS VLMs generate textual change descriptions but do not predict dense masks. \textbf{Additional experimental results on large images, out-of-domain dataset, qualitative interpretability, and parameter sensitivity are reported in \textit{Appendix~\ref{sec:appendix_experiments}}}.

\begin{table*}[!t]
\centering
\begin{tabular}{lcccccc}
\Xhline{1.5pt}
Model                  & BLEU-1 & ROUGE-1 & METEOR & Reranker & GPT-Judge & Human  \\ \Xhline{1.5pt}
Qwen3-VL-8B-Instruct   & 3.82      & 23.06      & 11.08      & 27.76          & 47.78          & 42.25          \\
Qwen3-VL-32B-Instruct  & 11.90     & 29.78      & 15.42      & 32.41          & 52.91          & 51.10          \\
Qwen3-VL-235B-Instruct & 5.25      & 23.76      & 11.77      & 56.20          & 64.53          & 60.30          \\ \hline
RSUniVLM              & 10.85     & 28.95      & 18.96      & 53.27          & 61.42          & 54.65          \\
Teochat                & 8.72      & 20.14      & 13.27      & 32.11          & 45.39          & 36.40          \\
EarthDial              & 7.13      & 24.43      & 17.84      & 51.10          & 59.83          & 51.80          \\ \hline
\textbf{QUAKE-CD}      & \textbf{49.30}      & \textbf{52.21}      & \textbf{40.24}      & \textbf{71.26} & \textbf{78.61} & \textbf{73.50} \\ \Xhline{1.5pt}
\end{tabular}
\caption{Semantic evaluation on the reasoning trace of model outputs. \textit{Human} annotators are given only the bi-temporal image pair and the model output.}
\label{tab:reasoning}
\end{table*}

\subsection{Dense Change Mask Generation}
\label{sec:exp_main}

For each prediction, we decode the extracted \textit{QUAKE} span into a binary mask with $\mathcal{D}$ and report Precision, Recall, and F1 under accumulated pixel statistics and per-image means. As shown in Table~\ref{tab:change_detection}, \textsc{QUAKE-CD} attains specialist-level mask quality while retaining a unified autoregressive interface for dense prediction and grounded reasoning. Its accumulated F1 of $78.31\%$ surpasses BIT by $1.37\%$ and remains within $3.13\%$ of the strongest specialist model, ChangeFormer. Under per-image evaluation, \textsc{QUAKE-CD} reaches $73.18\%$ F1 and outperforms all three specialist models. It further achieves the highest recall among all methods under both evaluation protocols. Against pixel-level VLMs, \textsc{QUAKE-CD} surpasses PixelLM by $51.04\%$ in accumulated F1 and LISA-7B by $51.27\%$ in per-image F1. Existing VLM baselines exhibit a pronounced recall--precision imbalance: they recover broad change regions but introduce extensive false positives. Even the strongest VLM baseline reaches only $50.12\%$ accumulated precision and $48.76\%$ per-image precision. Text4Seg exemplifies this regime, attaining $84.07\%$ accumulated recall but only $14.52\%$ precision and thus $24.76\%$ F1: its flat patch-grid serialization assigns uniform resolution across the scene and cannot resolve the small, fragmented objects characteristic of change detection. In contrast, \textsc{QUAKE-CD} raises precision to $70.99\%$ and $68.77\%$ while preserving strong recall at $87.31\%$ and $84.09\%$. This advantage stems from the selective allocation of quadtree tokens around change boundaries, which suppresses spurious regions without erasing fine-grained structures. RSUniVLM adopts flat text-as-mask formulation while accepting the bi-temporal pair natively. Its persistent recall--precision imbalance isolates the \textsc{QUAKE-CD} gain to hierarchical quadtree encoding rather than to native bi-temporal input.

\textsc{QUAKE-CD} also achieves a favorable balance between accuracy and efficiency. Its throughput reaches $74.27$ pairs/s, at least $1.85\times$ that of decoder-based VLMs, as native text generation avoids the overhead of a segmentation decoder. Its throughput remains close to flat serialization methods, including Text4Seg at $86.54$ pairs/s and RSUniVLM at $82.30$ pairs/s, despite the additional sequential structure introduced by the hierarchical \textit{QUAKE} syntax. In return, \textsc{QUAKE-CD} improves accumulated F1 over Text4Seg and RSUniVLM by $53.55\%$ and $15.97\%$, respectively, establishing a stronger trade-off between throughput and spatial fidelity. The detailed throughput setup is provided in Appendix~\ref{sec:appendix_hparams}.

\subsection{Reasoning Quality Evaluation}
\label{sec:exp_reasoning}

Since mask metrics cannot verify whether textual claims stem from bi-temporal evidence, we evaluate the generated \texttt{<think>} and \texttt{<answer>} trajectories. Although we report BLEU-1, ROUGE-1, and METEOR for lexical overlap, Table~\ref{tab:reasoning} shows that generic VLMs score abnormally low under $n$-gram metrics, which penalize paraphrasing and verbose reasoning rather than semantic correctness. We therefore complement them with two semantics-aware automatic evaluators and a human-rating protocol. \textbf{Reranker Score} (Qwen3-Reranker-4B) assesses semantic fidelity, completeness, and structural coherence. \textbf{GPT-Judge Accuracy} (GPT-5.5-2026-04-23) quantifies semantic consistency against human-verified references (prompt templates in Figure~\ref{fig:reranker_prompt} and \ref{fig:gpt_judge_prompt}). \textbf{Human Score} averages acceptance rates from two independent annotators over $1{,}000$ randomly sampled outputs assessed for factual consistency and localization correctness, reaching $87.5\%$ inter-annotator agreement.

Table~\ref{tab:reasoning} disentangles scaling from grounded bi-temporal supervision. Under traditional metrics, \textsc{QUAKE-CD} outperforms all baselines, achieving $49.30$ BLEU-1 and $40.24$ METEOR. On semantics-aware metrics, scaling Qwen3-VL from 8B to 235B raises the reranker score from $27.76$ to $56.20$ and GPT-judge accuracy from $47.78\%$ to $64.53\%$, as greater capacity improves fluency and coarse semantic plausibility. Yet scale alone does not close the grounding gap: \textsc{QUAKE-CD} exceeds the same-8B baseline by $31.25\%$ in human score and still leads the 235B model by $13.20\%$. RSUniVLM also trails \textsc{QUAKE-CD} by $18.85\%$ in human score, indicating that its change-captioning supervision is insufficient for capturing the key bi-temporal evidence underlying each change. The agreement across lexical, semantic, and human evaluators indicates that the gain is robust rather than an artifact of any single judge.

\begin{figure}[!t]
    \centering
    \includegraphics[width=0.79\columnwidth]{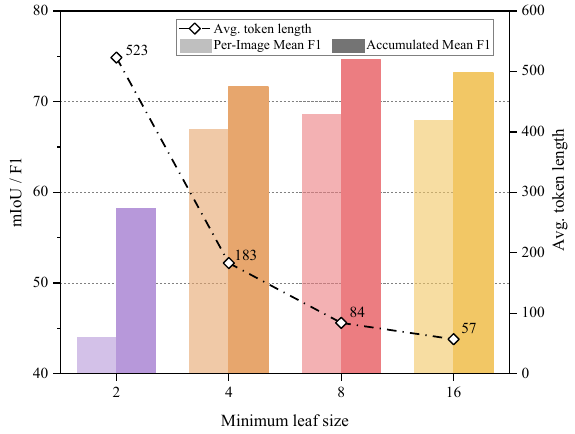}
    \caption{Ablation studies on the influence of the minimum leaf size $\ell$.}
    \label{fig:ablation_1}
\end{figure}

\begin{figure}[!ht]
    \centering
    \includegraphics[width=0.79\columnwidth]{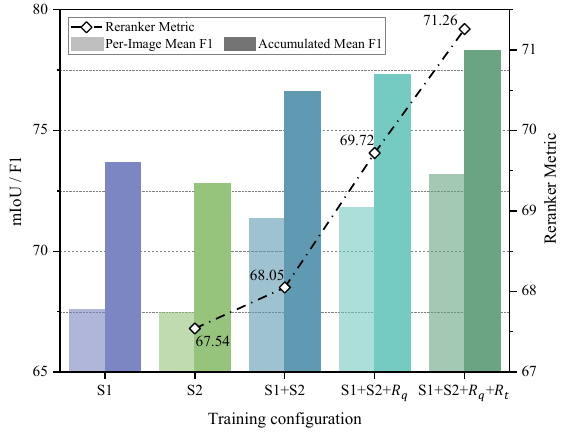}
    \caption{Ablation studies on the influence of curriculum stages and reward components.}
    \label{fig:ablation_2}
\end{figure}

\begin{figure*}[!t]
    \centering
    \includegraphics[width=0.89\linewidth]{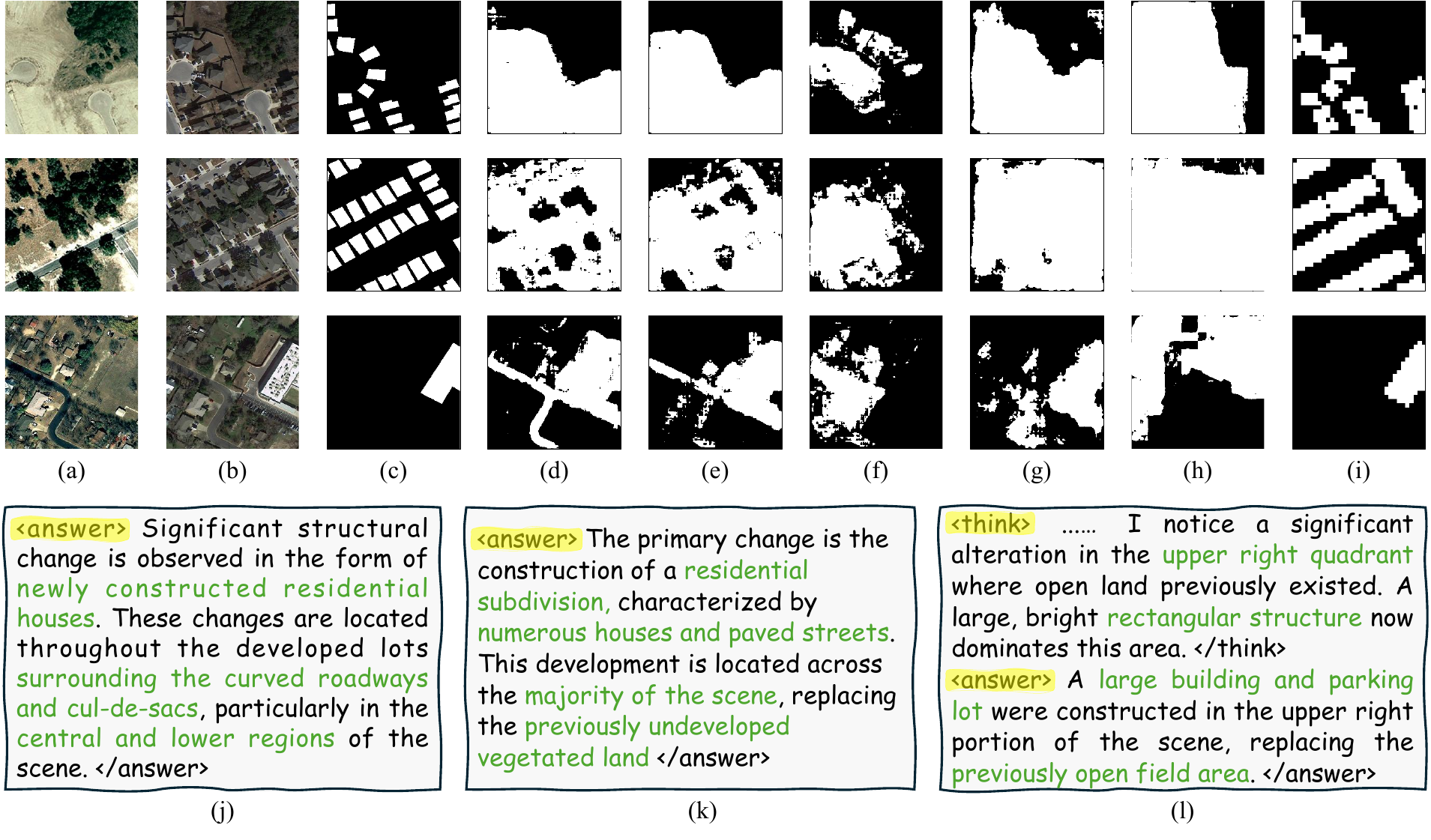}
    \caption{Qualitative results of representative decoder-based pixel-level VLMs and QUAKE-CD. (a): pre-change image. (b): post-change image. (c): ground truth. (d)-(i): results of LISA-7B, LISA-llama2-13B, GLaMM, PixelLM, Text4Seg, and QUAKE-CD. (j)–(l): textual change descriptions generated by QUAKE-CD.}
    \label{fig:case}
\end{figure*}

\subsection{Ablation Studies}
\label{sec:exp_ablation}

\noindent\textbf{Minimum Leaf Size.}
Figure~\ref{fig:ablation_1} varies the minimum leaf size $\ell$, which controls the trade-off between spatial resolution and sequence complexity, under fixed SFT settings. Reducing $\ell$ from $16$ to $2$ increases the average sequence length from $57$ to $523$ tokens, yet mask quality does not improve monotonically. F1 reaches its maximum at $\ell{=}8$ and drops sharply at $\ell{=}2$. This degradation reflects a learnability bottleneck rather than a representational limit: although finer leaves can encode sharper boundaries, long recursive sequences increase grammar violations and dimension mismatches, weakening both SFT supervision and RL rollouts. At the opposite extreme, $\ell{=}16$ reduces the token budget but quantizes each $16{\times}16$ patch into a single bit, which erodes boundary precision. Thus, $\ell{=}8$ provides the strongest operating point, preserving fine-grained structure while keeping autoregressive decoding and policy optimization tractable.

\noindent\textbf{Curriculum and Reward Components.}
Figure~\ref{fig:ablation_2} evaluates five training regimes: S1, S2, S1+S2, S1+S2+$R_q$, and S1+S2+$R_q$+$R_t$. The S1 regime---supervising only the \texttt{<seg>} block---serves as the \textit{QUAKE-only} variant that isolates the hierarchical representation from CoT supervision, while S1+S2 layers CoT atop the same syntax. First, the two-stage curriculum separates syntax acquisition from temporal reasoning: applying Stage~2 directly to the base VLM yields weaker per-image F1 due to interference between grammar learning and rationale generation, while initializing Stage-2 from the Stage-1 checkpoint improves both accumulated and per-image F1 by roughly $4\%$ over either single-stage variant. Notably, the S1-only mask quality already exceeds every flat-serialization baseline in Table~\ref{tab:change_detection}, attributing a substantial share of the dense-prediction gain to the QUAKE representation itself. Second, adding \textit{Quadtree Reward} raises accumulated F1 by about $1\%$ and the reranker score from $68.05$ to $69.72$, showing that grammar-gated spatial RL supplies a useful optimization signal beyond imitation learning. Third, the full dual-reward model reaches $71.26$ reranker score with further gains in accumulated F1, confirming that \textit{ThinkAnswer Reward} strengthens textual faithfulness without compromising the spatial accuracy preserved by \textit{Quadtree Reward}, and, in turn, provides a stronger semantic guidance for \textit{QUAKE} generation. Overall, the monotonic gains confirm that hierarchical encoding, staged supervision, and grammar-gated dual RL act synergistically across distinct failure modes.

\subsection{Qualitative Analysis}
\label{sec:exp_qualitative}

Figure~\ref{fig:case} presents representative cases where \textsc{QUAKE-CD} accurately resolves dense small-building construction requiring fine-grained bi-temporal comparison. External-decoder baselines and Text4Seg instead follow the high-recall pattern of Section~\ref{sec:exp_main}, over-predicting broad regions in dense scenes and hallucinating large changed areas in single-building cases where they miss the pre- and post-change structural correspondence. In contrast, \textsc{QUAKE-CD}'s textual descriptions remain grounded, correctly specifying change type, location, and extent.

\section{Conclusion}
\label{sec:conclusion}

We introduced \textsc{QUAKE-CD}, a framework that formulates dense remote-sensing change detection as structured autoregressive generation under a syntax-verifiable mask grammar. Its quadtree mask language converts binary change maps into compact, grammar-constrained token sequences that can be decoded, checked, and optimized within autoregressive VLMs. With paired spatial supervision, grounded reasoning traces, staged curriculum learning, and grammar-gated dual-reward RL, \textsc{QUAKE-CD} aligns dense mask generation with faithful bi-temporal reasoning. Experiments show clear gains over decoder-based and flat text-as-mask VLM baselines, while demonstrating that grounded spatial supervision improves reasoning more effectively than generic model scaling. These results position structured spatial languages as a practical foundation for accurate, interpretable, and structurally checkable dense prediction in VLMs.


\bibliography{aaai2027}

@article{qwen3-vl,
  title={Qwen3-vl technical report},
  author={Bai, Shuai and Cai, Yuxuan and Chen, Ruizhe and Chen, Keqin and Chen, Xionghui and Cheng, Zesen and Deng, Lianghao and Ding, Wei and Gao, Chang and Ge, Chunjiang and others},
  journal={arXiv preprint arXiv:2511.21631},
  year={2025}
}

@article{kimi-k2.5,
  title={Kimi K2.5: Visual Agentic Intelligence},
  author={Team, Kimi and Bai, Tongtong and Bai, Yifan and Bao, Yiping and Cai, SH and Cao, Yuan and Charles, Y and Che, HS and Chen, Cheng and Chen, Guanduo and others},
  journal={arXiv preprint arXiv:2602.02276},
  year={2026}
}

@article{vg-review,
  title={Towards visual grounding: A survey},
  author={Xiao, Linhui and Yang, Xiaoshan and Lan, Xiangyuan and Wang, Yaowei and Xu, Changsheng},
  journal={IEEE Transactions on Pattern Analysis and Machine Intelligence},
  year={2025},
  publisher={IEEE}
}

@article{res-review,
  title={Multimodal referring segmentation: A survey},
  author={Ding, Henghui and Tang, Song and He, Shuting and Liu, Chang and Wu, Zuxuan and Jiang, Yu-Gang},
  journal={arXiv preprint arXiv:2508.00265},
  year={2025}
}

@article{cd-review,
  title={Change detection techniques for remote sensing applications: A survey},
  author={Asokan, Anju and Anitha, JJESI},
  journal={Earth Science Informatics},
  volume={12},
  number={2},
  pages={143--160},
  year={2019},
  publisher={Springer}
}

@inproceedings{text4seg,
  title={Text4seg: Reimagining image segmentation as text generation},
  author={Lan, Mengcheng and Chen, Chaofeng and Zhou, Yue and Xu, Jiaxing and Ke, Yiping and Wang, Xinjiang and Feng, Litong and Zhang, Wei},
  booktitle={International Conference on Learning Representations},
  volume={2025},
  pages={1634--1661},
  year={2025}
}

@article{text4seg++,
  title={Text4seg++: Advancing image segmentation via generative language modeling},
  author={Lan, Mengcheng and Chen, Chaofeng and Xu, Jiaxing and Li, Zongrui and Ke, Yiping and Jiang, Xudong and Yu, Yingchen and Zhao, Yunqing and Bai, Song},
  journal={IEEE Transactions on Pattern Analysis and Machine Intelligence},
  year={2026},
  publisher={IEEE}
}

@inproceedings{changechat,
  title={Changechat: An interactive model for remote sensing change analysis via multimodal instruction tuning},
  author={Deng, Pei and Zhou, Wenqian and Wu, Hanlin},
  booktitle={ICASSP 2025-2025 IEEE International Conference on Acoustics, Speech and Signal Processing (ICASSP)},
  pages={1--5},
  year={2025},
  organization={IEEE}
}

@inproceedings{btcchat,
  title={BTCChat: Advancing remote sensing bi-temporal change captioning with multimodal large language model},
  author={Li, Yujie and Xu, Wenjia and Zhang, Yuanben and Wei, Zhiwei and Peng, Mugen},
  booktitle={ICASSP 2026-2026 IEEE International Conference on Acoustics, Speech and Signal Processing (ICASSP)},
  pages={4596--4600},
  year={2026},
  organization={IEEE}
}

@inproceedings{lisa,
  title={Lisa: Reasoning segmentation via large language model},
  author={Lai, Xin and Tian, Zhuotao and Chen, Yukang and Li, Yanwei and Yuan, Yuhui and Liu, Shu and Jia, Jiaya},
  booktitle={Proceedings of the IEEE/CVF conference on computer vision and pattern recognition},
  pages={9579--9589},
  year={2024}
}

@inproceedings{pixellm,
  title={Pixellm: Pixel reasoning with large multimodal model},
  author={Ren, Zhongwei and Huang, Zhicheng and Wei, Yunchao and Zhao, Yao and Fu, Dongmei and Feng, Jiashi and Jin, Xiaojie},
  booktitle={Proceedings of the IEEE/CVF Conference on Computer Vision and Pattern Recognition},
  pages={26374--26383},
  year={2024}
}

@inproceedings{glamm,
  title={Glamm: Pixel grounding large multimodal model},
  author={Rasheed, Hanoona and Maaz, Muhammad and Shaji, Sahal and Shaker, Abdelrahman and Khan, Salman and Cholakkal, Hisham and Anwer, Rao M and Xing, Eric and Yang, Ming-Hsuan and Khan, Fahad S},
  booktitle={Proceedings of the IEEE/CVF Conference on Computer Vision and Pattern Recognition},
  pages={13009--13018},
  year={2024}
}

@article{segearth-r1,
  title={Segearth-r1: Geospatial pixel reasoning via large language model},
  author={Li, Kaiyu and Xin, Zepeng and Pang, Li and Pang, Chao and Deng, Yupeng and Yao, Jing and Xia, Guisong and Meng, Deyu and Wang, Zhi and Cao, Xiangyong},
  journal={arXiv preprint arXiv:2504.09644},
  year={2025}
}

@article{grasp,
  title={GRASP: Guided Region-Aware Sparse Prompting for Adapting MLLMs to Remote Sensing},
  author={Sun, Qigan and Zhang, Chaoning and Zhang, Jianwei and Wang, Xudong and Xie, Jiehui and Zheng, Pengcheng and Wang, Haoyu and Lee, Sungyoung and Tai, Chi-lok Andy and Yang, Yang and others},
  journal={arXiv preprint arXiv:2601.17089},
  year={2026}
}

@inproceedings{remotereasoner,
  title={Remotereasoner: Towards unifying geospatial reasoning workflow},
  author={Yao, Liang and Liu, Fan and Lu, Hongbo and Zhang, Chuanyi and Min, Rui and Xu, Shengxiang and Di, Shimin and Peng, Pai},
  booktitle={Proceedings of the AAAI Conference on Artificial Intelligence},
  volume={40},
  pages={11883--11891},
  year={2026}
}

@article{rsthinker,
  title={Towards Faithful Reasoning in Remote Sensing: A Perceptually-Grounded GeoSpatial Chain-of-Thought for Vision-Language Models},
  author={Liu, Jiaqi and Sun, Lang and Fu, Ronghao and Yang, Bo},
  journal={arXiv preprint arXiv:2509.22221},
  year={2025}
}

@inproceedings{spatial-mm,
  title={An empirical analysis on spatial reasoning capabilities of large multimodal models},
  author={Shiri, Fatemeh and Guo, Xiao-Yu and Far, Mona Golestan and Yu, Xin and Haf, Reza and Li, Yuan-Fang},
  booktitle={Proceedings of the 2024 Conference on Empirical Methods in Natural Language Processing},
  pages={21440--21455},
  year={2024}
}

@article{grammar-aligned-decoding,
  title={Grammar-aligned decoding},
  author={Park, Kanghee and Wang, Jiayu and Berg-Kirkpatrick, Taylor and Polikarpova, Nadia and D'Antoni, Loris},
  journal={Advances in Neural Information Processing Systems},
  volume={37},
  pages={24547--24568},
  year={2024}
}

@article{grpo,
  title={Deepseek-r1: Incentivizing reasoning capability in llms via reinforcement learning},
  author={Guo, Daya and Yang, Dejian and Zhang, Haowei and Song, Junxiao and Wang, Peiyi and Zhu, Qihao and Xu, Runxin and Zhang, Ruoyu and Ma, Shirong and Bi, Xiao and others},
  journal={arXiv preprint arXiv:2501.12948},
  year={2025}
}

@article{changevg,
  title={Towards comprehensive interactive change understanding in remote sensing: A large-scale dataset and dual-granularity enhanced VLM},
  author={Xue, Junxiao and Deng, Quan and Wu, Xuecheng and Yao, Kelu and Yin, Xinyi and Yu, Fei and Zhou, Wei and Zhong, Yanfei and Liu, Yang and Yang, Dingkang},
  journal={IEEE Transactions on Geoscience and Remote Sensing},
  year={2026},
  publisher={IEEE}
}

@inproceedings{cure,
  title={Measuring and improving chain-of-thought reasoning in vision-language models},
  author={Chen, Yangyi and Sikka, Karan and Cogswell, Michael and Ji, Heng and Divakaran, Ajay},
  booktitle={Proceedings of the 2024 Conference of the North American Chapter of the Association for Computational Linguistics: Human Language Technologies (Volume 1: Long Papers)},
  pages={192--210},
  year={2024}
}

@article{sysu-cd,
  title={A deeply supervised attention metric-based network and an open aerial image dataset for remote sensing change detection},
  author={Shi, Qian and Liu, Mengxi and Li, Shengchen and Liu, Xiaoping and Wang, Fei and Zhang, Liangpei},
  journal={IEEE transactions on geoscience and remote sensing},
  volume={60},
  pages={1--16},
  year={2021},
  publisher={IEEE}
}

@article{levir-cd,
  title={A spatial-temporal attention-based method and a new dataset for remote sensing image change detection},
  author={Chen, Hao and Shi, Zhenwei},
  journal={Remote sensing},
  volume={12},
  number={10},
  pages={1662},
  year={2020},
  publisher={MDPI}
}

@article{gemini,
  title={Gemini: a family of highly capable multimodal models},
  author={Team, Gemini and Anil, Rohan and Borgeaud, Sebastian and Alayrac, Jean-Baptiste and Yu, Jiahui and Soricut, Radu and Schalkwyk, Johan and Dai, Andrew M and Hauth, Anja and Millican, Katie and others},
  journal={arXiv preprint arXiv:2312.11805},
  year={2023}
}

@article{tversky,
  title={Features of similarity.},
  author={Tversky, Amos},
  journal={Psychological review},
  volume={84},
  number={4},
  pages={327},
  year={1977},
  publisher={American Psychological Association}
}

@misc{ms-swift,
      title={SWIFT:A Scalable lightWeight Infrastructure for Fine-Tuning},
      author={Yuze Zhao and Jintao Huang and Jinghan Hu and Xingjun Wang and Yunlin Mao and Daoze Zhang and Zeyinzi Jiang and Zhikai Wu and Baole Ai and Ang Wang and Wenmeng Zhou and Yingda Chen},
      year={2024},
      eprint={2408.05517},
      archivePrefix={arXiv},
      primaryClass={cs.CL},
      url={https://arxiv.org/abs/2408.05517},
}

@inproceedings{vllm,
  title={Efficient Memory Management for Large Language Model Serving with PagedAttention},
  author={Woosuk Kwon and Zhuohan Li and Siyuan Zhuang and Ying Sheng and Lianmin Zheng and Cody Hao Yu and Joseph E. Gonzalez and Hao Zhang and Ion Stoica},
  booktitle={Proceedings of the ACM SIGOPS 29th Symposium on Operating Systems Principles},
  year={2023}
}

@article{levir-cc,
  title={Remote sensing image change captioning with dual-branch transformers: A new method and a large scale dataset},
  author={Liu, Chenyang and Zhao, Rui and Chen, Hao and Zou, Zhengxia and Shi, Zhenwei},
  journal={IEEE Transactions on Geoscience and Remote Sensing},
  volume={60},
  pages={1--20},
  year={2022},
  publisher={IEEE}
}

@article{dubai-cc,
  title={Change captioning: A new paradigm for multitemporal remote sensing image analysis},
  author={Hoxha, Genc and Chouaf, Seloua and Melgani, Farid and Smara, Youcef},
  journal={IEEE Transactions on Geoscience and Remote Sensing},
  volume={60},
  pages={1--14},
  year={2022},
  publisher={IEEE}
}

@article{rsunivlm,
  title={RSUniVLM: A unified vision-language model for remote sensing via Granularity-oriented MoE},
  author={Liu, Xu and Lian, Zhouhui},
  journal={Pattern Recognition},
  pages={113717},
  year={2026},
  publisher={Elsevier}
}

@article{geollava,
  title={Geollava: Efficient fine-tuned vision-language models for temporal change detection in remote sensing},
  author={Elgendy, Hosam and Sharshar, Ahmed and Aboeitta, Ahmed and Ashraf, Yasser and Guizani, Mohsen},
  journal={arXiv preprint arXiv:2410.19552},
  year={2024}
}

@article{vilacd,
  title={ViLaCD-R1: A Vision-Language Framework for Semantic Change Detection in Remote Sensing},
  author={Ma, Xingwei and Feng, Shiyang and Zhang, Bo and Wang, Bin},
  journal={arXiv preprint arXiv:2512.23244},
  year={2025}
}

@article{remotevar,
  title={RemoteVAR: Autoregressive Visual Modeling for Remote Sensing Change Detection},
  author={Korkmaz, Yilmaz and Patel, Vishal M},
  journal={arXiv preprint arXiv:2601.11898},
  year={2026}
}

@article{decode_the_delta,
  title={Decoding the Delta: Unifying Remote Sensing Change Detection and Understanding with Multimodal Large Language Models},
  author={Li, Xiaohe and Li, Jiahao and Zhang, Kaixin and Fang, Yuqiang and Lin, Leilei and Wang, Hong and Wu, Haohua and Fan, Zide},
  journal={arXiv preprint arXiv:2604.14044},
  year={2026}
}

@inproceedings{semantic-cd,
  title={Semantic-CD: Remote sensing image semantic change detection towards open-vocabulary setting},
  author={Zhu, Yongshuo and Li, Lu and Chen, Keyan and Liu, Chenyang and Zhou, Fugen and Shi, Zhenwei},
  booktitle={IGARSS 2025-2025 IEEE International Geoscience and Remote Sensing Symposium},
  pages={6388--6392},
  year={2025},
  organization={IEEE}
}

@inproceedings{calico,
  title={Calico: Part-focused semantic co-segmentation with large vision-language models},
  author={Nguyen, Kiet A and Juvekar, Adheesh and Yu, Tianjiao and Wahed, Muntasir and Lourentzou, Ismini},
  booktitle={Proceedings of the Computer Vision and Pattern Recognition Conference},
  pages={4550--4561},
  year={2025}
}

@inproceedings{mc-bench,
  title={Mc-bench: A benchmark for multi-context visual grounding in the era of mllms},
  author={Xu, Yunqiu and Zhu, Linchao and Yang, Yi},
  booktitle={Proceedings of the IEEE/CVF International Conference on Computer Vision},
  pages={17675--17687},
  year={2025}
}

@inproceedings{sam,
  title={Segment anything},
  author={Kirillov, Alexander and Mintun, Eric and Ravi, Nikhila and Mao, Hanzi and Rolland, Chloe and Gustafson, Laura and Xiao, Tete and Whitehead, Spencer and Berg, Alexander C and Lo, Wan-Yen and others},
  booktitle={Proceedings of the IEEE/CVF international conference on computer vision},
  pages={4015--4026},
  year={2023}
}

@inproceedings{teochat,
  title={Teochat: A large vision-language assistant for temporal earth observation data},
  author={Irvin, Jeremy and Liu, Emily and Chen, Joyce and Dormoy, Ines and Kim, Jinyoung and Khanna, Samar and Zheng, Zhuo and Ermon, Stefano},
  booktitle={International Conference on Learning Representations},
  volume={2025},
  pages={68883--68911},
  year={2025}
}

@inproceedings{earthdial,
  title={Earthdial: Turning multi-sensory earth observations to interactive dialogues},
  author={Soni, Sagar and Dudhane, Akshay and Debary, Hiyam and Fiaz, Mustansar and Munir, Muhammad Akhtar and Danish, Muhammad Sohail and Fraccaro, Paolo and Watson, Campbell D and Klein, Levente J and Khan, Fahad Shahbaz and others},
  booktitle={Proceedings of the Computer Vision and Pattern Recognition Conference},
  pages={14303--14313},
  year={2025}
}

@article{lora,
  title={Lora: Low-rank adaptation of large language models.},
  author={Hu, Edward J and Shen, Yelong and Wallis, Phillip and Allen-Zhu, Zeyuan and Li, Yuanzhi and Wang, Shean and Wang, Liang and Chen, Weizhu and others},
  journal={Iclr},
  volume={1},
  number={2},
  pages={3},
  year={2022}
}

@article{bit,
  title={Remote sensing image change detection with transformers},
  author={Chen, Hao and Qi, Zipeng and Shi, Zhenwei},
  journal={IEEE Transactions on Geoscience and Remote Sensing},
  volume={60},
  pages={1--14},
  year={2021},
  publisher={IEEE}
}

@inproceedings{changeformer,
  title={A transformer-based siamese network for change detection},
  author={Bandara, Wele Gedara Chaminda and Patel, Vishal M},
  booktitle={IGARSS 2022-2022 IEEE International Geoscience and Remote Sensing Symposium},
  pages={207--210},
  year={2022},
  organization={IEEE}
}

@article{changeclip,
  title={ChangeCLIP: Remote sensing change detection with multimodal vision-language representation learning},
  author={Dong, Sijun and Wang, Libo and Du, Bo and Meng, Xiaoliang},
  journal={ISPRS Journal of Photogrammetry and Remote Sensing},
  volume={208},
  pages={53--69},
  year={2024},
  publisher={Elsevier}
}

@ARTICLE{whu-building,
  author={Ji, Shunping and Wei, Shiqing and Lu, Meng},
  journal={IEEE Transactions on Geoscience and Remote Sensing}, 
  title={Fully Convolutional Networks for Multisource Building Extraction From an Open Aerial and Satellite Imagery Data Set}, 
  year={2019},
  volume={57},
  number={1},
  pages={574-586},
  doi={10.1109/TGRS.2018.2858817}}

\clearpage
\setcounter{secnumdepth}{1}
\appendix
\section{Quadtree Mask Encoding Details}
\label{sec:appendix_quadtree}

This appendix extends the informal description in Section~\ref{sec:quadtree_encoding} with a complete formal specification of $\mathcal{L}_Q$, its deterministic encoder $E$, and decoder $\mathcal{D}$. The specification renders $\mathcal{L}_Q$ a syntactically and structurally checkable mask language whose well-formedness is decidable in linear time, independent of the model's semantic predictions.

\subsection{Formal Grammar of {$\mathcal{L}_Q$}}

The alphabet of $\mathcal{L}_Q$ extends the base VLM vocabulary with 19 special tokens, supplemented by standard separators and numerals:
\begin{itemize}
    \item \textbf{Structural delimiters}: \texttt{<seg>}, \texttt{</seg>}
    \item \textbf{Uniform tile tokens}: \texttt{<BG>}, \texttt{<FG>}
    \item \textbf{Quadtree node tokens}: 15 nodes $\{\texttt{<Q0001>},\ldots,\texttt{<Q1111>}\}$; the all-zero pattern \texttt{<Q0000>} is strictly excluded, since an inactive quadrant is already absorbed by its parent node's occupancy bit.
    \item \textbf{Separators and Modifiers}: \texttt{|} (column delimiter), \texttt{\textbackslash n} (row delimiter), and \texttt{*} (run-length encoding prefix).
\end{itemize}

The syntax of $\mathcal{L}_Q$ is governed by a context-free grammar defined by the following production rules:
\begin{align}
    \text{Sequence} &\to \texttt{<seg>}\; \text{Grid}\; \texttt{</seg>} \\
    \text{Grid}     &\to \text{Row}\; (\texttt{\textbackslash n}\; \text{Row})^* \\
    \text{Row}      &\to \text{Tile}\; (\texttt{|}\; \text{Tile})^* \nonumber\\
    \text{Tile}     &\to \texttt{<BG>} \mid \texttt{<FG>} \mid \texttt{<BG>}*n \mid \texttt{<FG>}*n \nonumber\\
                    &\quad \mid \texttt{<Q}b_1 b_2 b_3 b_4\texttt{>}\; \text{Children}(b_1 b_2 b_3 b_4, d)
\end{align}
Here, $\text{Children}(b_1 b_2 b_3 b_4, d)$ recursively expands every active quadrant with $b_i = 1$ in Z-order (preorder) traversal: top-left, top-right, bottom-left, bottom-right. Let $d$ denote the current recursion depth, bounded by $d_{\max} = \lceil\log_2(32/\ell)\rceil$. Upon reaching $d_{\max}$, the quadrant collapses to a base $\text{Tile}$ of side length $\ell$.

Grammar validity ensures format compliance before decoding. A sequence is valid if and only if: (1) exactly one \texttt{<seg>} wrapper encloses the grid; (2) the expanded row and column counts match the expected grid dimensions $H/32 \times W/32$; (3) every run-length count satisfies $n \ge 2$ and does not exceed the remaining tiles in its row; (4) quadtree recursion depth never exceeds $d_{\max}$; and (5) the number of recursively expanded children matches the sum of active bits $\sum b_i$ in the corresponding \texttt{<Q}$abcd$\texttt{>} node.

\subsection{Encoding Algorithm {$E$}}

Algorithm~\ref{alg:encode} specifies the deterministic encoding procedure. Given a binary mask $M \in \{0,1\}^{H \times W}$, a fixed tile size $T = 32$, and a minimum leaf size $\ell$, the encoder partitions $M$ into an $H/T \times W/T$ grid and recursively subdivides each heterogeneous tile until the minimum leaf size $\ell$ is reached.

\begin{algorithm}[t]
    \centering
    \small
    \begin{algorithmic}[1]
        \Require Binary mask $M \in \{0,1\}^{H \times W}$, tile size $T = 32$, min leaf size $\ell = 8$
        \Ensure Quadtree sequence $\mathbf{y}_{\text{seg}} \in \mathcal{L}_Q$
        \State Partition $M$ into grid $\mathcal{G}$ of $H/T \times W/T$ tiles of size $T \times T$
        \State Initialize sequence buffer $S \gets$ \texttt{<seg>}
        \For{each row $r \in \mathcal{G}$}
            \For{each tile $t \in r$}
                \If{$t$ is uniform background ($\forall_{i,j}\, t[i,j] = 0$)}
                    \State Emit \texttt{<BG>}
                \ElsIf{$t$ is uniform foreground ($\forall_{i,j}\, t[i,j] = 1$)}
                    \State Emit \texttt{<FG>}
                \Else
                    \State Emit \Call{EncodeNode}{$t$, $T$, $\ell$, depth $= 0$}
                \EndIf
                \State Emit column delimiter \texttt{|} if $t$ is not the last tile in $r$
            \EndFor
            \State Apply RLE: collapse consecutive identical uniform tokens to \texttt{<BG>}*$n$ or \texttt{<FG>}*$n$
            \State Append row separator \texttt{\textbackslash n}
        \EndFor
        \State Append \texttt{</seg>}
        \State \Return $S$

        \Function{EncodeNode}{$B$, size, $\ell$, depth}
            \State $d_{\max} \gets \lceil\log_2(\text{size} / \ell)\rceil$
            \If{depth $= d_{\max}$}
                \State \Return \texttt{<FG>} if quadrant contains any foreground pixel else \texttt{<BG>} \Comment{recall-preserving leaf rule}
            \EndIf
            \State $h \gets \text{size} / 2$
            \State $b_1 b_2 b_3 b_4 \gets$ occupancy bits of quadrants TL, TR, BL, BR
            \If{$b_1 = b_2 = b_3 = b_4 = 0$} \Comment{unreachable; defensive guard}
                \State \Return empty string
            \EndIf
            \State Emit \texttt{<Q}$b_1 b_2 b_3 b_4$\texttt{>}
            \For{each quadrant $q_i$ with $b_i = 1$ in TL, TR, BL, BR order}
                \State \Call{EncodeNode}{$B[q_i]$, $h$, $\ell$, depth $+ 1$}
            \EndFor
        \EndFunction
    \end{algorithmic}
    \caption{Quadtree Mask Encoding}
    \label{alg:encode}
\end{algorithm}

The encoding is \emph{recall-preserving} by construction: if $M[i,j] = 1$, then $\mathcal{D}(E(M))[i,j] = 1$. A single foreground pixel within any leaf forces that leaf's occupancy bit to 1, and the decoder renders every active leaf as fully foreground. Precision loss therefore originates exclusively from boundary leaves of size $\ell$, where mixed background pixels are unavoidably promoted to foreground. For $\ell = 8$, the maximum false-positive expansion per boundary leaf is $64$ pixels. This controlled precision--compression trade-off strictly eliminates false-negative boundary omissions, accounting for the bounded precision drop.

Total complexity is $O(HW)$: each pixel is visited once during partitioning, and recursive subdivisions operate on disjoint subregions.

\subsection{Decoding Algorithm {$\mathcal{D}$}}

Decoding reconstructs a binary mask $\hat{M}$ from a quadtree sequence $\mathbf{y}_{\mathrm{seg}} \in \mathcal{L}_Q$ as a purely symbolic, non-parametric mapping, eliminating neural overhead during both inference and RL reward computation.

Upon extracting the \texttt{<seg>...\text{</seg>}} body, the decoder splits the grid by row separators, expands RLE prefixes, and validates the reconstructed sequence dimensions. Invalid sequences trigger categorical safety fallbacks:
\begin{itemize}
    \item \textbf{Grammar Failure:} Missing wrappers, unrecognized node tokens, or malformed RLE prefixes yield an all-zero fallback mask.
    \item \textbf{Dimension Mismatch:} If expanded dimensions diverge from $H/32 \times W/32$, the decoder applies pad/crop alignment to recover a usable mask; the corresponding reward-side discount is specified in Appendix~\ref{sec:appendix_dual_reward}.
    \item \textbf{Unexpected EOF:} Severely truncated sequences that break during token parsing default to the all-zero mask.
    \item \textbf{Depth Overflow:} Recursion exceeding $d_{\max}$ triggers an immediate fallback to a uniform $\ell \times \ell$ leaf-level approximation.
\end{itemize}

For valid sequences, tile tokens are rendered deterministically. \texttt{<BG>} and \texttt{<FG>} generate $32 \times 32$ zero or one blocks, respectively. A \texttt{<Q}$abcd$\texttt{>} node recursively renders only the quadrants whose occupancy bits $b_i = 1$, while inactive quadrants default to background. The rendered spatial blocks are concatenated in raster-scan order into the final dense mask $\hat{M} \in \{0,1\}^{H \times W}$.

\subsection{Encoding Fidelity and Compression Statistics}
\label{sec:appendix_encoding_fidelity}

Table~\ref{tab:appendix_encoding_fidelity} quantifies the round-trip algorithmic fidelity of encoding followed by decoding across the QUAKE-CoT dataset. Foreground recall strictly achieves $100.0\%$ at both accumulated and per-image levels, directly validating the claim of preserving all target foreground regions. Accumulated F1 of $91.2\%$ and per-image F1 of $88.0\%$ confirm that boundary-quantization precision loss remains strictly bounded under $\ell{=}8$. The decoded masks therefore serve as faithful supervision targets for SFT and as reliable reward signals for RL.

\begin{table}[t]
\centering
\begin{tabular}{lcc}
\Xhline{1.2pt}
Metric & Accumulated (\%) & Per-Image Mean (\%) \\
\Xhline{1.2pt}
IoU & 83.9 & 80.2 \\
F1  & 91.2 & 88.0 \\
Precision & 83.9 & 80.2 \\
Recall & 100.0 & 100.0 \\
\Xhline{1.2pt}
\end{tabular}%
\caption{Round-trip encoding fidelity of $E$ and $\mathcal{D}$ on the QUAKE-CoT training set.}
\label{tab:appendix_encoding_fidelity}
\end{table}

Table~\ref{tab:appendix_leaf_ablation} ablates the minimum leaf size $\ell$. Setting $\ell{=}2$ inflates sequence length, while $\ell{=}16$ shortens sequences at the cost of spatial resolution and a higher per-leaf false-positive rate. We adopt $\ell{=}8$ as the operating point that balances compression against precision.

\begin{table}[!t]
\centering
\begin{tabular}{lcccc}
\Xhline{1.2pt}
$\ell$ & IoU & Accum.\ F1 & Per-Image F1 & Avg.\ Tokens \\
\Xhline{1.2pt}
2  & 97.2 & 98.6 & 97.9 & 520.7 \\
4  & 91.2 & 95.3 & 93.8 & 124.7 \\
8  & 83.9 & 91.2 & 88.0 & 78.3 \\
16 & 71.4 & 83.2 & 79.7 & 52.1 \\
\Xhline{1.2pt}
\end{tabular}%
\caption{Encoding fidelity by minimum leaf size $\ell$ on the QUAKE-CoT training set. Smaller leaves improve precision; larger leaves reduce sequence length.}
\label{tab:appendix_leaf_ablation}
\end{table}

\section{QUAKE-CoT Dataset Details}
\label{sec:appendix_dataset}

QUAKE-CoT pairs pixel-level change masks with chain-of-thought reasoning traces, supplying the linguistic supervision absent from purely discriminative change-detection benchmarks. We build upon established bi-temporal remote sensing benchmarks, augmenting them with hierarchical mask sequences and teacher-distilled reasoning traces.

\subsection{Data Sources and Curation}

We source raw image pairs and binary change masks from three large-scale change detection benchmarks: SYSU-CD \citep{sysu-cd}, LEVIR-CD and LEVIR-CD+ \citep{levir-cd}. Table~\ref{tab:appendix_source_datasets} summarizes the characteristics of these source datasets. SYSU-CD covers diverse urban change types including buildings, roads, and vegetation, whereas the LEVIR series targets building additions and removals.

\begin{table*}[t]
\centering
\begin{tabular}{lccc}
\Xhline{1.2pt}
Source & Resolution & Main change type & \# Samples \\
\Xhline{1.2pt}
SYSU-CD & $256 \times 256$ & Urban changes & $20{,}000$ \\
LEVIR-CD & $1,024 \times 1,024$ & Building changes & $637$ \\
LEVIR-CD+ & $1,024 \times 1,024$ & Building and subtle changes & $985$ \\
\Xhline{1.2pt}
\end{tabular}%
\caption{Source datasets used to construct QUAKE-CoT. All original images were cropped or resized to $256 \times 256$ spatial resolution.}
\label{tab:appendix_source_datasets}
\end{table*}

The curation process involves three stages:

\paragraph{Spatial Quantization.}
SYSU-CD natively provides $256{\times}256$ pairs and is consumed as-is. LEVIR-CD and LEVIR-CD+ ($1024{\times}1024$) are partitioned into 16 non-overlapping $256{\times}256$ tiles per pair via raster scan. Each mask $M$ is then encoded into $\mathbf{y}_{\mathrm{seg}}$ via Algorithm~\ref{alg:encode} with $T{=}32$ and $\ell{=}8$. No positive--negative balancing is applied at this stage; sample-level filtering is deferred to the validation step described below.

\paragraph{Reasoning Distillation.}
We employ \textit{Gemini-3.1-Pro-Thinking-Preview}~\cite{gemini} as the teacher, conditioning generation on the bi-temporal pair $(I_1, I_2)$ and the ground-truth mask $M$. The prompt in Figure~\ref{fig:qa_prompt} enforces two core constraints. First, the trace must reference only the two temporal images and pursue reasoning-oriented comparison---articulating how differences should be analyzed and which categories of change are relevant---rather than producing yes/no judgements. Second, the teacher must explicitly discount appearance shifts caused by season, phenology, sun angle, shadows, haze, or global brightness/contrast, focusing instead on structural change such as new or removed objects, altered built footprints, land-cover conversion, and layout reconfiguration. We issue $55{,}740$ teacher calls in total; $9{,}788$ ($17.6\%$) require re-invocation due to API timeouts or format violations, and the affected samples are retained only after a successful re-call. Reasoning traces tokenized with the Qwen3-VL tokenizer average $84$ tokens, with median $70$ and range $31$--$303$.

\paragraph{Validation.}
Four graduate students with backgrounds in remote sensing, computer vision, or computer science cross-verify all $N_{\mathrm{trace}}$ distilled traces against the corresponding bi-temporal pair and ground-truth mask, revising obvious factual inconsistencies. Traces judged factually consistent account for $98.7\%$ of the corpus, while the remaining $1.3\%$ are revised by the annotators. We therefore define the trace-validation rate as
\begin{equation}
    A_{\mathrm{trace}}
    = \frac{N_{\mathrm{consistent}}}{N_{\mathrm{trace}}}
    = 98.7\%.
\end{equation}
We then apply three sample-level filters: (i) traces that remain factually inconsistent with the mask after revision are discarded; (ii) tiles whose change consists solely of sub-tile-scale boundary noise---small fragmented edge variations deemed too ambiguous for chain-of-thought training---are removed; (iii) to mitigate the heavy class imbalance toward unchanged tiles, we retain a uniformly random $20\%$ of fully-empty-mask tiles. Filtering yields the final samples documented in Table~\ref{tab:appendix_splits}.

\paragraph{Human Annotation and Agreement.}
All human assessments were conducted by the same four graduate annotators using a common rubric for factual consistency and spatial grounding. For the reported Human Score, we evaluated $N=1{,}000$ randomly sampled outputs for each of the $M=7$ evaluated models. Let $h_{m,i}^{(a)}\in\{0,1\}$ denote the acceptance judgment assigned by annotator $a\in\{1,2\}$ to the $i$-th output of model $m$. The acceptance rate of annotator $a$ for model $m$ is
\begin{equation}
    A_m^{(a)} = \frac{1}{N}\sum_{i=1}^{N} h_{m,i}^{(a)},
\end{equation}
and the reported Human Score for model $m$ averages the two annotators' acceptance rates:
\begin{equation}
    H_m = \frac{1}{2}\left(A_m^{(1)} + A_m^{(2)}\right)
    = \frac{1}{2N}\sum_{i=1}^{N}\sum_{a=1}^{2} h_{m,i}^{(a)}.
\end{equation}
Inter-annotator agreement is computed as exact label agreement over all $M N$ evaluated outputs:
\begin{equation}
    A_{\mathrm{inter}}
    = \frac{1}{M N}\sum_{m=1}^{M}\sum_{i=1}^{N}
    \mathbb{1}\!\left[h_{m,i}^{(1)} = h_{m,i}^{(2)}\right]
    = 87.5\%.
\end{equation}
To rule out the concern that the trained model merely replays teacher templates rather than grounding its reasoning in visual evidence, two of the same annotators additionally audit a stratified $500$-sample subset of \textsc{QUAKE-CD} inference outputs, manually verifying that each rationale's spatial claims---location, change type, and approximate extent---agree with the model's own predicted mask. The rationale--mask agreement rate is defined as
\begin{equation}
    A_{\mathrm{rm}}
    = \frac{N_{\mathrm{mask\text{-}consistent}}}{500}
    = 97.2\%,
\end{equation}
indicating that the textual reasoning is internally consistent with the spatial prediction rather than decoupled boilerplate.

\subsection{Data Partitioning}

The dataset is partitioned into training and testing sets following the splits defined in the source benchmarks, enabling direct comparison with prior change-detection methods. Table~\ref{tab:appendix_splits} details the final data splits. The test split is held out for the final evaluation of both the SFT baselines and the GRPO-trained QUAKE-CD model.

\begin{table}[t]
\centering
\begin{tabular}{lcc}
\Xhline{1.2pt}
Split & Size & Purpose \\
\Xhline{1.2pt}
Train & $22{,}007$ & Stage 1 \& Stage 2 \\
Train & $5{,}502$ & RM training \& GRPO \\
Test & $9{,}301$ & Final evaluation \\
\Xhline{1.2pt}
\end{tabular}%
\caption{QUAKE-CoT data partition. The dataset is divided into a training split and a held-out test split, maintaining the original dataset distributions.}
\label{tab:appendix_splits}
\end{table}

\subsection{Dataset Statistics}
\label{sec:appendix_dataset_statistics}

The final QUAKE-CoT corpus comprises $36{,}810$ aligned $\langle$image pair, mask, reasoning trace$\rangle$ triples drawn from the three source benchmarks. Reasoning traces total approximately $3.09$M Qwen3-VL tokens, averaging $60.4$ words ($84$ tokens) per trace---roughly an order of magnitude longer than the single-sentence captions in LEVIR-CC~\cite{levir-cc} ($7.99$ words) and DUBAI-CC~\cite{dubai-cc} ($7.35$ words). QUAKE-CoT is also the only corpus among the compared datasets to pair these reasoning traces with deterministically decodable dense mask sequences.

\begin{table*}[t]
\centering
\begin{tabular}{lccccc}
\Xhline{1.2pt}
Dataset & Primary task & Caption / QA & Mask & Reasoning Trace & Avg. Word Count \\
\Xhline{1.2pt}
LEVIR-CC & Change captioning & \checkmark & -- & -- & 7.99 \\
DUBAI-CC & Change captioning & \checkmark & -- & -- & 7.35 \\
ChangeChat & Interactive change understanding & \checkmark & -- & Partial & -- \\
SYSU-CD & Binary change detection & -- & \checkmark & -- & -- \\
LEVIR-CD & Binary change detection & -- & \checkmark & -- & -- \\
LEVIR-CD+ & Binary change detection & -- & \checkmark & -- & -- \\ \hline
\textbf{QUAKE-CoT} & Structured dense change generation & \checkmark & \checkmark & \checkmark & 60.38 \\
\Xhline{1.2pt}
\end{tabular}%
\caption{Comparison between QUAKE-CoT and representative remote-sensing change-language and change-detection datasets. Captioning datasets provide language supervision but do not provide dense binary masks for autoregressive mask generation. Conventional change-detection datasets provide dense masks but no reasoning traces. QUAKE-CoT pairs bi-temporal images with teacher-distilled reasoning traces and deterministic quadtree mask sequences.}
\label{tab:appendix_dataset_comparison}
\end{table*}

\subsection{Comparison with Existing Benchmarks}

Table~\ref{tab:appendix_dataset_comparison} compares QUAKE-CoT with representative datasets across change detection and change captioning paradigms. Standard discriminative benchmarks such as LEVIR-CD and SYSU-CD provide only binary masks, leaving language supervision to downstream users. Existing change-captioning datasets like LEVIR-CC~\cite{levir-cc} and DUBAI-CC~\cite{dubai-cc} attach a single short caption per pair, which seldom encodes the spatial localization or step-wise comparison needed for chain-of-thought training. QUAKE-CoT couples each pair with a multi-step reasoning trace and a deterministically decodable dense mask, supporting joint optimization of language and spatial grounding within a single autoregressive objective.

\begin{figure}[!t]
    \centering
    \includegraphics[width=\columnwidth]{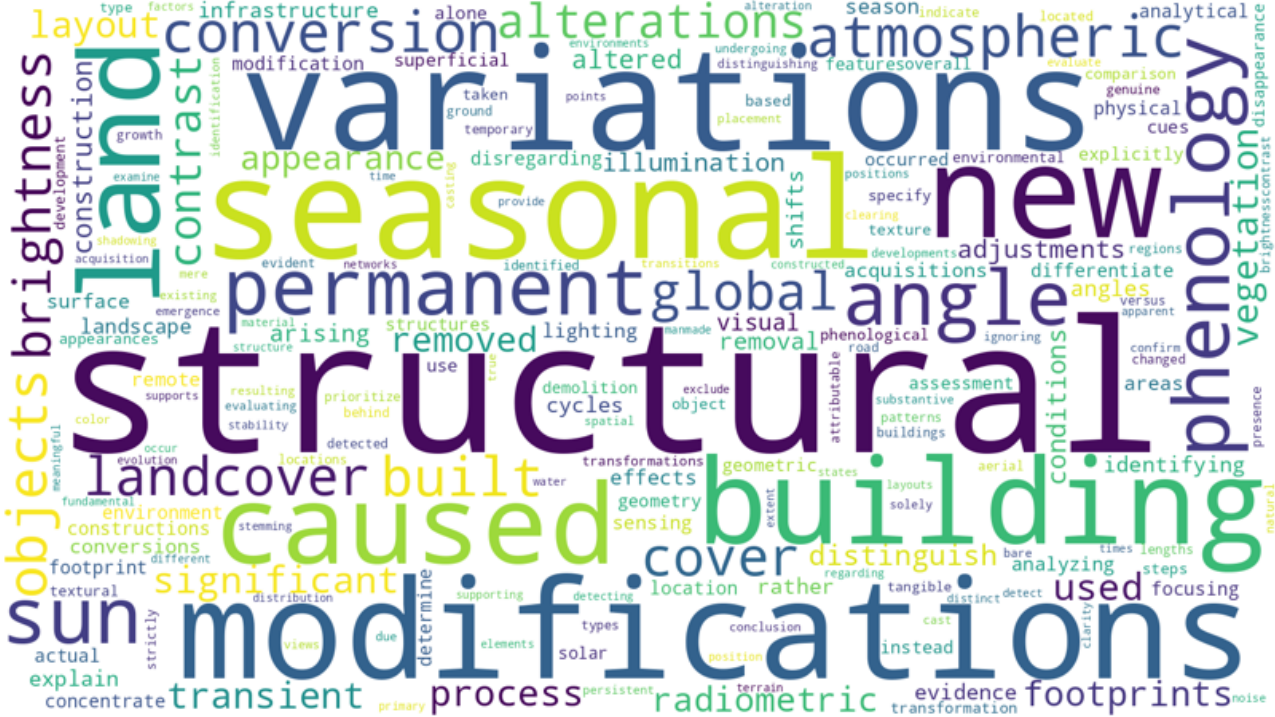}
    \caption{Word cloud visualization of the QUAKE-CoT dataset's textual content.}
    \label{fig:wordcloud}
\end{figure}

Figure~\ref{fig:wordcloud} visualizes the lexical distribution of QUAKE-CoT's reasoning traces. The vocabulary is dominated by structural objects (\textit{building}, \textit{landcover}) and change descriptors (\textit{new}, \textit{modification}, \textit{conversion}), reflecting the spatial--temporal vocabulary required for bi-temporal reasoning rather than the surface-level descriptors typical of single-sentence captioning.

\subsection{Licenses and Limitations}
\label{sec:appendix_dataset_limitations}

QUAKE-CoT is released under the licenses of its underlying source datasets (SYSU-CD, LEVIR-CD, and LEVIR-CD+); we redistribute the bi-temporal imagery together with the derived quadtree mask sequences and reasoning traces under those terms.

We highlight three limitations relevant to downstream use. First, reasoning traces are distilled from \textit{Gemini-3.1-Pro-Thinking-Preview}; biases or systematic blind spots of the teacher---for example, preferred phrasings, residual reliance on global appearance cues, or culturally specific descriptors---may propagate into the trained policy despite the human cross-verification pass. Second, the source benchmarks emphasize built-environment changes, so building-level additions and removals are over-represented relative to natural changes such as vegetation succession, water-body fluctuation, or land-cover conversion at fine granularity. Third, the validation filter removes tiles whose annotated change is dominated by sub-tile-scale boundary noise, which trims a tail of legitimate but very small changes; QUAKE-CoT therefore underweights the high-frequency micro-change regime and should not be the sole benchmark for evaluating sub-pixel or thin-linear change detection.

\section{Reward Construction Details}
\label{sec:appendix_rm}

This section describes the training data and objective for the ThinkAnswer Reward $R_\phi$, which is trained in a dedicated RM stage and subsequently consumed by the GRPO RL stage. The training set combines policy rollouts scored by IoU, one Gemini-synthesized hard negative per query, and the ground-truth trace, assembled into strict pairwise preferences over a five-tier hierarchy.

\subsection{Rollout Generation and IoU Scoring}

For each prompt $\mathbf{x}$ in the $20\%$ RL subset of the QUAKE-CoT dataset, we independently sample $K=10$ responses $\{y_1,\ldots,y_K\}\sim \pi_{\theta_{\mathrm{SFT}}}(\cdot\mid\mathbf{x})$ from the curriculum-SFT policy. Each response $y_i$ is parsed to extract its quadtree mask sequence $y_{\text{seg}}^{(i)}$. The sequence is subsequently processed by the deterministic decoder $\mathcal{D}$ to yield the binary mask prediction $\hat{M}^{(i)} = \mathcal{D}(y_{\text{seg}}^{(i)})$. 

We evaluate the structural quality of every rollout by computing its Intersection over Union (IoU) with the ground-truth mask $M$:
\begin{equation}
    \text{IoU}(y_i) = \frac{|\hat{M}^{(i)} \cap M|}{|\hat{M}^{(i)} \cup M|}
\end{equation}
Generations that fail $\mathcal{L}_Q$ grammar matching---missing \texttt{<seg>} wrapper, unrecognized tokens, or malformed RLE prefixes---are assigned $\mathrm{IoU}(y_i)=0$; pad/crop fallbacks triggered by dimension mismatch still receive a normally computed IoU.

\subsection{Hard-Negative Generation}

For each query we synthesize one fluent-but-contradictory hard negative $y_{\text{neg}}$, forcing $R_\phi$ to discriminate factual fidelity from surface fluency rather than rewarding stylistic eloquence.

\paragraph{Generator and Prompt.}
The generator is \textit{Gemini-3.1-Pro-Thinking-Preview}~\cite{gemini}, queried via API and conditioned on the bi-temporal pair $(I_1, I_2)$ together with the ground-truth trace $y_{\text{gt}}$. Following the system prompt in Figure~\ref{fig:hard_negative_prompt}, the generator is instructed to inject exactly one factual conflict into the rewritten trace, drawn from three categories: fabricating a change that did not occur, displacing the change to an incorrect location, or asserting no-change when change is present.

\paragraph{Format Constraints.}
To ensure $y_{\text{neg}}$ is difficult to distinguish from genuine outputs, the prompt enforces three constraints: (i) the linguistic length is bounded between $0.8\times$ and $1.2\times$ of $|y_{\text{gt}}|$ in tokens; (ii) the original \texttt{<think>...\text{</think>}\text{<answer>}...\text{</answer>}} structure is preserved verbatim; (iii) giveaway tokens such as ``mask'' or ``ground truth'' are forbidden.

$y_{\text{neg}}$ is therefore placed at the bottom of the preference hierarchy defined below.

\subsection{Preference Pair Construction}

From the $K=10$ rollouts we select three samples spanning the IoU spectrum:
\begin{equation}
    y_a = \arg\max_{y_i}\,\mathrm{IoU}(y_i),
\end{equation}
\begin{equation}
    y_b = \arg\min_{y_i}\,\mathrm{IoU}(y_i),
\end{equation}
\begin{multline}
    y_c = \arg\max_{y_i\notin\{y_a,y_b\}} \bigl|\mathrm{IoU}(y_a)-\mathrm{IoU}(y_i)\bigr| \\
    \cdot \bigl|\mathrm{IoU}(y_i)-\mathrm{IoU}(y_b)\bigr|.
\end{multline}
Here $y_a$ and $y_b$ are the highest- and lowest-IoU rollouts, while $y_c$ is the intermediate rollout whose IoU sits farthest from both endpoints, measured by the product of absolute IoU gaps.

A preference comparison between any two candidates $(y_i, y_j)$ is considered valid only if their performance discrepancy is significant enough to warrant a discernible semantic difference. We enforce this through a margin check threshold $\tau = 0.1$:
\begin{equation}
    y_i \succ y_j \iff \text{IoU}(y_i) - \text{IoU}(y_j) \ge \tau
\end{equation}

By incorporating the ground-truth sequence $y_{\text{gt}}$ (considered the absolute upper bound) and the fabricated hard negative $y_{\text{neg}}$ (the absolute lower bound), we establish a stringent, multi-tiered preference hierarchy:
\begin{equation}
    y_{\text{gt}} \succ y_a \succ y_c \succ y_b \succ y_{\text{neg}}
\end{equation}

From this hierarchy, we extract the preference set
\begin{equation}
    \mathcal{D}_{\mathrm{pref}} = \bigl\{(y_c, y_r)\,\bigm|\, y_c \succ y_r\bigr\},
\end{equation}
instantiated by pairs such as $(y_{\text{gt}}, y_{\text{neg}})$, $(y_b, y_{\text{neg}})$, $(y_a, y_c)$, and $(y_c, y_b)$, provided the $\tau$ gap condition is satisfied for rollout comparisons. These pairs are aggregated to train the Qwen3-VL-8B-Instruct reward model. The neural reward model is optimized using the following objective:
\begin{equation}
  \begin{aligned}
  \mathcal{L}_{\mathrm{RM}}
  = -\mathbb{E}_{\mathcal{D}_{\mathrm{pref}}}\big[
  &\log \sigma(R_\phi(y_c) - R_\phi(y_r)) \\
  &- \lambda (R_\phi(y_c) + R_\phi(y_r))^2
  \big],
  \end{aligned}
  \label{eq:appendix_rm}
\end{equation}
The resulting preference set anchors $R_\phi$ to spatial fidelity rather than surface phrasing.

\subsection{Dual-Reward Specification}
\label{sec:appendix_dual_reward}

During the GRPO RL stage, the total reward combines a rule-based $R_q$ with the learned $R_t$:
\begin{equation}
    R(\mathbf{x}, y) = (1-\mu)\,R_q(y;M) + \mu\,R_t(y;\mathbf{x}),
\end{equation}
with $\mu = 0.1$. The rule-based component validates grammar and spatial accuracy; the learned component scores the bi-temporal reasoning trace.

\paragraph{Quadtree Reward Formulation ($R_q$).} 
The Quadtree Reward deterministically validates format compliance and spatial accuracy:
\begin{equation}
    R_q = \text{clip}(R_{\text{struct}} + R_{\text{content}},\; 0.0,\; 1.0)
\end{equation}
If the generation fails $\mathcal{L}_Q$ parsing---missing \texttt{<seg>} wrapper, unrecognized tokens, a forbidden \texttt{<Q0000>} node, an RLE count below $2$, or a child count not matching $\sum_i b_i$ for any \texttt{<Q}$abcd$\texttt{>} node---grammar gating forces $R_q = 0.0$. Otherwise:
\begin{itemize}
    \item \textbf{Structure Score ($R_{\text{struct}}$):} The decoded grid is compared against the expected dimensions $(H_e, W_e)$ via row-wise error rates and an exact-alignment bonus:
    \begin{equation}
        \epsilon_h = \min\!\bigl(1,\, |H_p - H_e|/H_e\bigr),
    \end{equation}
    \begin{equation}
        \epsilon_w = \frac{1}{H_e}\sum_{i=1}^{H_e} \min\!\bigl(1,\, |W_i - W_e|/W_e\bigr),
    \end{equation}
    \begin{multline}
        R_{\text{struct}} = 0.25\,\max\!\bigl(0,\,1 - \tfrac{1}{2}(\epsilon_h+\epsilon_w)\bigr) \\
        + 0.05\cdot\mathbb{1}\bigl[H_p = H_e \wedge \forall i\,W_i = W_e\bigr],
    \end{multline}
    where $H_p$ is the decoded row count and $W_i$ the logical width of row $i$ after RLE expansion.
    \item \textbf{Content Score ($R_{\text{content}}$):} The decoded binary mask $\hat{M}$ is scored against the ground truth $M$ via the Tversky index $\mathcal{T}_{\alpha,\beta}$ with $(\alpha,\beta)=(0.3,0.7)$, which weights false negatives more strongly than false positives. Writing $\mathrm{TP}=|M\cap\hat{M}|$, $\mathrm{FP}=|\hat{M}\setminus M|$, and $\mathrm{FN}=|M\setminus\hat{M}|$,
    \begin{equation}
        \mathcal{T}_{\alpha,\beta}(M,\hat{M}) = \frac{\mathrm{TP}}{\mathrm{TP} + \alpha\,\mathrm{FP} + \beta\,\mathrm{FN}}.
    \end{equation}
    The content score is non-trivial only when the ground-truth mask contains foreground: when $|M|=0$, an exact-empty prediction earns a residual reward of $0.2$ to prevent collapse to all-background outputs, while any false-positive prediction yields $0$. For active masks ($|M|>0$):
    \begin{equation}
    \begin{aligned}
    R_{\text{content}} = \, &0.70\,\mathcal{T}_{\alpha,\beta}(M,\hat{M})\,s_{\text{align}} \\
    &+ 0.03\,\mathbb{1}[\mathcal{T}\ge 0.45 \wedge \text{strict}],
    \end{aligned}
    \end{equation}
    where $s_{\text{align}} = 1.0$ under strict decode and $s_{\text{align}} = 0.80$ when the pad/crop fallback was triggered. The additive $+0.03$ rewards strict-decoded predictions whose Tversky score clears the boundary-preservation threshold.
\end{itemize}

\begin{table*}[!ht]
\centering
\begin{tabular}{l|cccc}
\toprule
\textbf{Hyperparameter} & \textbf{Stage 1 (SFT)} & \textbf{Stage 2 (SFT)} & \textbf{RM Training} & \textbf{RL (GRPO)} \\
\midrule
Foundation Model & Qwen3-VL-8B & Stage 1 Checkpoint & Qwen3-VL-8B & Stage 2 Checkpoint \\
Training Epochs & $10$ & $5$ & $3$ & $3$ \\
Optimizer & AdamW & AdamW & AdamW & AdamW \\
Precision & BF16 & BF16 & BF16 & BF16 \\
Learning Rate & $2 \times 10^{-4}$ & $2 \times 10^{-4}$ & $1 \times 10^{-4}$ & $5 \times 10^{-6}$ \\
LR Scheduler & Cosine & Cosine & Cosine & Cosine \\
Warmup Ratio & $0.03$ & $0.03$ & $0.05$ & $0.03$ \\
Weight Decay & $0.01$ & $0.01$ & $0.01$ & $0.01$ \\
Hardware Setup & $8\times$ H200 & $8\times$ H200 & $4\times$ H200 & $8\times$ H200 \\
Batch Size (BS) per GPU & $8$ & $8$ & $4$ & $2$ \\
Gradient Accumulation Steps & $1$ & $1$ & $1$ & $1$ \\
Max Sequence Length & $4096$ & $4096$ & $2048$ & $1024$ (Completion) \\
\midrule
\multicolumn{5}{l}{\textit{LoRA Configuration}} \\
\midrule
LoRA Target Modules & Attention \& MLP & Attention \& MLP & Attention \& MLP & All Linear \\
LoRA Rank ($r$) & $64$ & $64$ & $64$ & $32$ \\
LoRA Alpha ($\alpha$) & $128$ & $128$ & $128$ & $64$ \\
LoRA Dropout & $0.05$ & $0.05$ & $0.05$ & $0.05$ \\
\midrule
\multicolumn{5}{l}{\textit{RL-Specific Configuration}} \\
\midrule
Rollouts per Prompt ($K$) & -- & -- & -- & $8$ \\
Temperature & -- & -- & -- & $0.7$ \\
KL Coefficient ($\beta$) & -- & -- & -- & $0.04$ \\
Reward Weights & -- & -- & -- & $1.0\ (R_q),\, 0.1\ (R_t)$ \\
\bottomrule
\end{tabular}%
\caption{Comprehensive hyperparameters for all training stages of \textsc{QUAKE-CD}. BS = batch size per device.}
\label{tab:appendix_hparams}
\end{table*}

\paragraph{ThinkAnswer Reward Formulation ($R_t$).} 
The reward model $R_\phi$ trained in the preceding RM stage scores the extracted \texttt{<think>...\text{</think>}\text{<answer>}...\text{</answer>}} segment of each rollout. To stabilise the reward scale across queries, we apply query-wise centred sigmoid normalisation over the group $\mathcal{S}_{\mathbf{x}}$ of $K=8$ rollouts sharing prompt $\mathbf{x}$:
\begin{equation}
    R_t(y_i) = \sigma\bigl( R_{\phi}(y_i \mid \mathbf{x}) - \bar{R}_{\phi}(\mathbf{x}) \bigr),
\end{equation}
\begin{equation}
    \bar{R}_{\phi}(\mathbf{x}) = \frac{1}{|\mathcal{S}_{\mathbf{x}}|} \sum_{y_j \in \mathcal{S}_{\mathbf{x}}} R_{\phi}(y_j \mid \mathbf{x}).
\end{equation}
The mean is computed across all $K$ rollouts of $\mathbf{x}$ aggregated over all data-parallel ranks, so every rank applies identical centring; this bounds $R_t \in (0,1)$ and yields zero-mean advantages within each query group.

\section{Comprehensive Hyperparameters}
\label{sec:appendix_hparams}

Table~\ref{tab:appendix_hparams} lists the full hyperparameter configuration for the two SFT stages, the reward model, and the GRPO rollout.

\noindent\textbf{Throughput Measurement Protocol.} Because Table~\ref{tab:change_detection} mixes specialist change detectors and VLMs, we define throughput at the application level rather than the token level. All methods are measured on a single NVIDIA H200 GPU with batch size $64$ under BF16 inference, and the reported numbers are wall-clock end-to-end throughput in pairs/s. For specialist detectors, this includes direct mask prediction and deterministic post-processing. For VLM baselines, it includes prompt encoding, autoregressive generation until the model's stopping criterion, and final mask extraction when applicable. For \textsc{QUAKE-CD}, this additionally includes generation of the \texttt{<seg>} span, followed by deterministic decoding of \texttt{<seg>} into the binary mask. No test-time adaptation, model parallelism, or extra refinement is used.

\section{More Experimental Results}
\label{sec:appendix_experiments}

\subsection{Oracle Performance Upper Bound}
\label{sec:appendix_oracle}
The mask representation imposes an upper bound independently of model generation. Table~\ref{tab:appendix_encoding_fidelity_comparison} reports the representation oracles of the hierarchical \textit{QUAKE} encoding and the flat text-as-mask encoding on the QUAKE-CoT test set. The flat oracle applies to Text4Seg and RSUniVLM, while the \textit{QUAKE} oracle applies to \textsc{QUAKE-CD}.

\begin{table*}[t]
\centering
\begin{tabular}{lcccccc}
\Xhline{1.2pt}
\multirow{2}{*}{Representation} & \multicolumn{3}{c}{Accumulated Mean} & \multicolumn{3}{c}{Per-Image Mean} \\ \cmidrule(lr){2-4}\cmidrule(lr){5-7}
 & Precision & Recall & F1 & Precision & Recall & F1 \\ \Xhline{1.2pt}
Flat text-as-mask & 68.04 & \textbf{100.00} & 80.97 & 65.23 & \textbf{100.00} & 78.98 \\
\textbf{QUAKE} & \textbf{84.03} & \textbf{100.00} & \textbf{91.32} & \textbf{81.43} & \textbf{100.00} & \textbf{88.68} \\ \Xhline{1.2pt}
\end{tabular}
\caption{Representation-oracle mask quality on the QUAKE-CoT test set. Each ground-truth mask is encoded with the corresponding mask representation and deterministically decoded without model generation. All metrics are reported as percentages.}
\label{tab:appendix_encoding_fidelity_comparison}
\end{table*}

\begin{table*}[!t]
\centering
\begin{tabular}{lcccccc}
\Xhline{1.2pt}
\multirow{2}{*}{Method} & \multicolumn{3}{c}{Accumulated Mean} & \multicolumn{3}{c}{Per-Image Mean} \\ \cmidrule(lr){2-4}\cmidrule(lr){5-7}
                & Precision & Recall & F1 & Precision & Recall & F1 \\ \Xhline{1.2pt}
Text4Seg        & 21.34     & 84.07  & 30.58 & 23.41     & 54.88  & 24.88 \\
RSUniVLM        & 73.67     & 82.45  & 76.99 & 74.74     & 81.23  & 77.17 \\ \hline
\textbf{QUAKE-CD} & \textbf{84.48} & \textbf{87.31} & \textbf{85.75} & \textbf{84.45} & \textbf{84.09} & \textbf{82.52} \\ \Xhline{1.2pt}
\end{tabular}%
\caption{End-to-end mask quality normalized by the corresponding representation oracle on the QUAKE-CoT test set. Each entry reports model performance divided by oracle performance.}
\label{tab:appendix_oracle_performance}
\end{table*}

\begin{table*}[!t]
\centering
\begin{tabular}{lcccccc}
\Xhline{1.2pt}
\multirow{2}{*}{Dataset} & \multicolumn{3}{c}{Accumulated Mean} & \multicolumn{3}{c}{Per-Image Mean} \\ \cmidrule(lr){2-4}\cmidrule(lr){5-7}
                & Precision & Recall & F1 & Precision & Recall & F1 \\ \Xhline{1.2pt}
LEVIR-CD        & 64.65 & 98.61 & 78.10 & 58.53 & 90.23 & 71.00 \\
LEVIR-CD+       & 55.79 & 85.82 & 67.62 & 57.23 & 78.68 & 66.26 \\ \Xhline{1.2pt}
\end{tabular}%
\caption{Change-mask performance under tiled inference on the original $1024\times1024$ test images. All values are percentages.}
\label{tab:appendix_high_resolution}
\end{table*}

Table~\ref{tab:appendix_encoding_fidelity_comparison} isolates the effect of mask representation. Both encodings preserve $100.00\%$ foreground recall, confirming that neither representation discards annotated change regions. Compared with flat text-as-mask encoding, \textit{QUAKE} increases accumulated precision from $68.04\%$ to $84.03\%$ and per-image precision from $65.23\%$ to $81.43\%$. These improvements raise accumulated and per-image F1 by $10.35\%$ and $9.70\%$, respectively. The results show that hierarchical refinement suppresses boundary over-coverage induced by uniform spatial quantization while preserving complete foreground coverage.

For metric $m \in \{\mathrm{Precision},\mathrm{Recall},\mathrm{F1}\}$ and evaluation regime $s \in \{\mathrm{acc},\mathrm{img}\}$, the fraction of oracle performance reached is
\begin{equation}
    \rho_{m}^{(s)}
    = \frac{m_{\mathrm{model}}^{(s)}}
           {m_{\mathrm{oracle}}^{(s)}} \times 100\%,
    \label{eq:appendix_oracle_fraction}
\end{equation}
where $m_{\mathrm{oracle}}^{(s)}$ is the upper bound obtained from representation round-trip fidelity and $m_{\mathrm{model}}^{(s)}$ is the end-to-end result in Table~\ref{tab:change_detection}. These ratios quantify the remaining model-generation gap after accounting for representation-specific quantization.

Table~\ref{tab:appendix_oracle_performance} shows that \textsc{QUAKE-CD} realizes the largest fraction of its representation oracle across all metrics. It reaches $85.75\%$ accumulated F1 and $82.52\%$ per-image F1, exceeding the strongest alternative by $8.76\%$ and $5.35\%$, respectively. Its normalized precision also remains above $84\%$ under both evaluation regimes, indicating substantially less degradation from autoregressive generation. Together, the two tables show that the gains of \textsc{QUAKE-CD} arise from both a more faithful spatial representation and a more effective realization of its attainable mask quality.

\subsection{Scalability to High-Resolution Imagery}
\label{sec:appendix_large_images}
Although our model is trained with $256\times256$ image pairs, practical remote sensing imagery often covers substantially larger spatial extents. We therefore evaluate its scalability on the original $1024\times1024$ images from LEVIR-CD and LEVIR-CD+, without high-resolution fine-tuning or architectural modification. Each co-registered bi-temporal pair is partitioned into a $4\times4$ grid of non-overlapping and spatially aligned $256\times256$ tile pairs. The model processes each pair independently at its native training resolution. The resulting \textit{QUAKE} predictions are decoded into binary tile masks, restored to their original coordinates, and stitched into a full-resolution $1024\times1024$ prediction for evaluation against the original ground truth. This protocol preserves local image detail and avoids the information loss caused by resizing the entire image to the training resolution. No overlapping inference, test-time adaptation, cross-tile refinement, or learned fusion is employed.

Table~\ref{tab:appendix_high_resolution} reports the resulting full-resolution performance. On the test pairs of LEVIR-CD, the model achieves a dataset-level precision of $64.65\%$ and an F1 score of $78.10\%$, with a recall of $98.61\%$. On the test pairs of LEVIR-CD+, it obtains a precision of $55.79\%$ and an F1 score of $67.62\%$, with a recall of $85.82\%$. The consistently high recall indicates that the model retains strong sensitivity to changed structures when applied to images with $16\times$ the training image area. The lower precision on LEVIR-CD+ suggests that the remaining errors are primarily attributable to conservative over-detection rather than missed changes. Overall, these results demonstrate that the proposed \textit{QUAKE} predictor can be effectively extended to megapixel-scale imagery through a simple tiled inference strategy while preserving the original training resolution.

\begin{figure*}[!t]
    \centering
    \includegraphics[width=\linewidth]{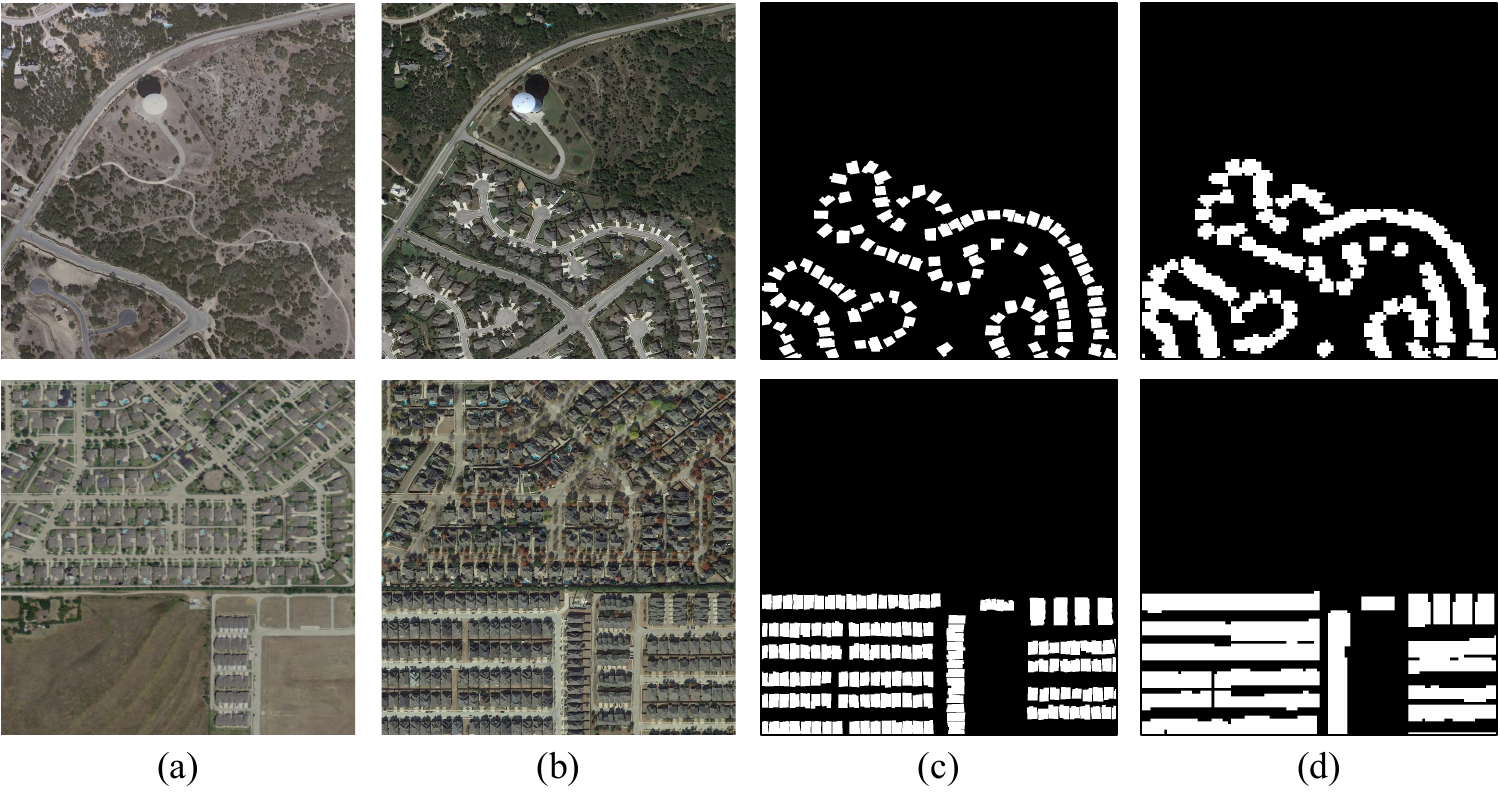}
    \caption{Qualitative results on original $1024\times1024$ image pairs under tiled inference. Each row presents one example: (a) pre-change image; (b) post-change image; (c) ground-truth change mask; (d) stitched prediction.}
    \label{fig:appendix_large_images}
\end{figure*}

Figure~\ref{fig:appendix_large_images} provides qualitative evidence consistent with these results. In the first example, the stitched prediction recovers the full extent of the newly developed curvilinear residential area, while merging nearby building footprints and expanding into some surrounding regions. In the second example, it captures the long rows of newly constructed buildings but represents several adjacent instances as contiguous bands. These errors explain the high recall and lower precision reported in Table~\ref{tab:appendix_high_resolution}. Despite independent tile processing, both predictions retain the global spatial arrangement of changes after stitching, confirming effective transfer from $256\times256$ training crops to $1024\times1024$ scenes.

\begin{table*}[!t]
\centering
\begin{tabular}{lcccccc}
\Xhline{1.2pt}
\multirow{2}{*}{Method} & \multicolumn{3}{c}{Accumulated Mean} & \multicolumn{3}{c}{Per-Image Mean} \\ \cmidrule(lr){2-4}\cmidrule(lr){5-7}
                & Precision & Recall & F1 & Precision & Recall & F1 \\ \Xhline{1.2pt}
\textcolor{gray}{BIT}          & \textcolor{gray}{72.93} & \textcolor{gray}{68.25} & \textcolor{gray}{68.54} & \textcolor{gray}{62.24} & \textcolor{gray}{60.44} & \textcolor{gray}{60.21} \\
\textcolor{gray}{ChangeFormer} & \textcolor{gray}{75.83} & \textcolor{gray}{71.04} & \textcolor{gray}{75.36} & \textcolor{gray}{71.30} & \textcolor{gray}{60.63} & \textcolor{gray}{65.29} \\
\textcolor{gray}{ChangeCLIP}   & \textcolor{gray}{76.41} & \textcolor{gray}{68.70} & \textcolor{gray}{73.09} & \textcolor{gray}{54.43} & \textcolor{gray}{45.76} & \textcolor{gray}{47.13} \\ \hline
LISA-7B         & 15.73 & 64.03 & 25.96 & 16.46 & 42.15 & 19.97 \\
LISA-llama2-13B & 14.88 & 68.52 & 22.99 & 17.62 & 41.71 & 21.15 \\
GLaMM           & 12.34 & 45.36 & 23.00 & 14.23 & 27.53 & 15.93 \\
PixelLM         & 16.02 & 65.63 & 26.75 & 17.72 & 42.73 & 20.87 \\ \hline
Text4Seg        & 14.65 & 74.41 & 22.26 & 12.94 & 47.80 & 19.18 \\
RSUniVLM        & 45.23 & 74.57 & 58.25 & 46.80 & 72.62 & 56.65 \\ \hline
\textsc{QUAKE-CD}                         & \textbf{65.36} & \textbf{78.40} & \textbf{71.29} & \textbf{64.90} & \textbf{73.88} & \textbf{69.10} \\
\textcolor{gray}{\textsc{QUAKE-CD} / Oracle} & \textcolor{gray}{77.84} & \textcolor{gray}{78.40} & \textcolor{gray}{78.10} & \textcolor{gray}{87.69} & \textcolor{gray}{73.88} & \textcolor{gray}{82.44} \\ \Xhline{1.2pt}
\end{tabular}%
\caption{Out-of-domain performance on WHU Building CD. \textsc{QUAKE-CD} is evaluated without specific fine-tuning. The second row reports end-to-end performance normalized by the corresponding representation oracle.}
\label{tab:appendix_whu}
\end{table*}

\subsection{Generalization on Out-of-Domain Dataset}
\label{sec:appendix_generalization}
All preceding evaluations use data drawn from QUAKE-CoT and therefore retain substantial overlap with its training distribution. We assess external validity on WHU Building CD \cite{whu-building}, which is excluded from data construction, supervised fine-tuning, and reinforcement learning. WHU has a ground sampling distance of $0.2$ m/pixel, compared with $0.3$ m/pixel for LEVIR-CD and $0.5$ m/pixel for SYSU-CD, and was acquired with a different sensor. This benchmark consequently introduces coupled shifts in spatial resolution, sensor characteristics, scene appearance, and object scale.

We evaluate the frozen \textsc{QUAKE-CD} model without specific fine-tuning or test-time adaptation. The native ground sampling distance is preserved, and each pair is processed using the same tiled inference and deterministic decoding protocol described in Section~\ref{sec:experiments}.

\begin{table*}[!t]
\centering
\begin{tabular}{lcccccc}
\Xhline{1.2pt}
\multirow{2}{*}{$(\alpha,\beta)$} & \multicolumn{3}{c}{Accumulated Mean} & \multicolumn{3}{c}{Per-Image Mean} \\ \cmidrule(lr){2-4}\cmidrule(lr){5-7}
                & Precision & Recall & F1 & Precision & Recall & F1 \\ \Xhline{1.2pt}
(0.1, 0.9)      & 70.82 & 86.68 & 77.95 & 68.41 & 83.04 & 72.63 \\
(0.2, 0.8)      & 70.24 & 86.89 & 77.68 & 68.19 & 83.24 & 72.51 \\
\textbf{(0.3, 0.7)} & 70.99 & \textbf{87.31} & \textbf{78.31} & \textbf{68.77} & \textbf{84.09} & \textbf{73.18} \\
(0.4, 0.6)      & \textbf{73.13} & 81.19 & 76.95 & 69.63 & 79.44 & 71.61 \\
(0.5, 0.5)      & 73.09 & 81.31 & 76.98 & 69.40 & 79.35 & 71.46 \\ \Xhline{1.2pt}
\end{tabular}
\caption{Sensitivity of the Tversky parameters in the Quadtree Reward.}
\label{tab:appendix_tversky_sensitivity}
\end{table*}

\begin{table*}[!t]
\centering
\begin{tabular}{lccccccc}
\Xhline{1.2pt}
\multirow{2}{*}{$\mu$} & \multicolumn{3}{c}{Accumulated Mean} & \multicolumn{3}{c}{Per-Image Mean} & Reranker \\ \cmidrule(lr){2-4}\cmidrule(lr){5-7}
                & Precision & Recall & F1 & Precision & Recall & F1 & (score) \\ \Xhline{1.2pt}
0.05            & 68.91 & 76.38 & 72.45 & 58.70 & 74.50 & 67.50 & 69.72 \\
\textbf{0.10}   & 70.99 & \textbf{87.31} & \textbf{78.31} & 68.77 & \textbf{84.09} & \textbf{73.18} & \textbf{71.26} \\
0.15            & 70.55 & 86.75 & 77.82 & 68.31 & 83.17 & 72.60 & 70.30 \\
0.20            & \textbf{73.09} & 81.31 & 76.98 & \textbf{69.40} & 79.35 & 71.46 & 68.05 \\ \Xhline{1.2pt}
\end{tabular}
\caption{Sensitivity of the reward trade-off coefficient in Eq.~(3).}
\label{tab:appendix_mu_sensitivity}
\end{table*}

\begin{table*}[!t]
\centering
\begin{tabular}{lccccccc}
\Xhline{1.2pt}
\multirow{2}{*}{KL Coefficient} & \multicolumn{3}{c}{Accumulated Mean} & \multicolumn{3}{c}{Per-Image Mean} & Reranker \\ \cmidrule(lr){2-4}\cmidrule(lr){5-7}
                & Precision & Recall & F1 & Precision & Recall & F1 & (score) \\ \Xhline{1.2pt}
\textbf{0.00}   & \textbf{72.78} & 81.53 & 76.90 & \textbf{69.44} & 79.67 & 71.58 & 64.85 \\
0.02            & 71.60 & 82.91 & 76.84 & 68.77 & 80.66 & 71.64 & 70.82 \\
\textbf{0.04}   & 70.99 & \textbf{87.31} & \textbf{78.31} & 68.77 & \textbf{84.09} & \textbf{73.18} & \textbf{71.26} \\
0.06            & 68.91 & 76.38 & 72.45 & 66.03 & 74.50 & 67.50 & 70.79 \\
0.08            & 69.25 & 76.31 & 72.61 & 66.12 & 74.46 & 67.48 & 64.09 \\ \Xhline{1.2pt}
\end{tabular}
\caption{Sensitivity of the GRPO KL coefficient. The Reranker column reports the average reranker score.}
\label{tab:appendix_kl_sensitivity}
\end{table*}

Table~\ref{tab:appendix_whu} reports the end-to-end mask quality of \textsc{QUAKE-CD} and the fraction of the corresponding representation oracle attained by the model. The oracle is obtained by encoding and deterministically decoding the WHU ground-truth masks with the \textit{QUAKE} representation. The normalized row follows Equation~\ref{eq:appendix_oracle_fraction} and divides each end-to-end metric by its oracle counterpart. Under accumulated evaluation, \textsc{QUAKE-CD} remains competitive with the specialist change detectors: its accumulated F1 of $71.29\%$ is within $4.07\%$ of the strongest specialist, ChangeFormer, while its accumulated recall of $78.40\%$ is the best in the table. More importantly, \textsc{QUAKE-CD} surpasses all specialist baselines in the per-image setting, reaching $69.10\%$ F1 versus $65.29\%$ for the strongest specialist, ChangeFormer. Among the VLM baselines, \textsc{QUAKE-CD} is the best by a clear margin; the strongest VLM baseline, RSUniVLM, reaches only $58.25\%$ accumulated F1 and $56.65\%$ per-image F1, leaving \textsc{QUAKE-CD} ahead by $13.04$ and $12.45$ points, respectively. Without adaptation, \textsc{QUAKE-CD} achieves accumulated and per-image F1 scores of $71.29\%$ and $69.10\%$. These results correspond to $78.10\%$ and $82.44\%$ of the respective oracle F1 scores. Recall remains higher than precision under both evaluation regimes, showing that the transferred model favors coverage of changed buildings over conservative boundary estimation. The results establish effective generalization across shifts in spatial resolution, sensor characteristics, and scene appearance.

\begin{figure*}[!t]
    \centering
    \includegraphics[width=0.95\linewidth]{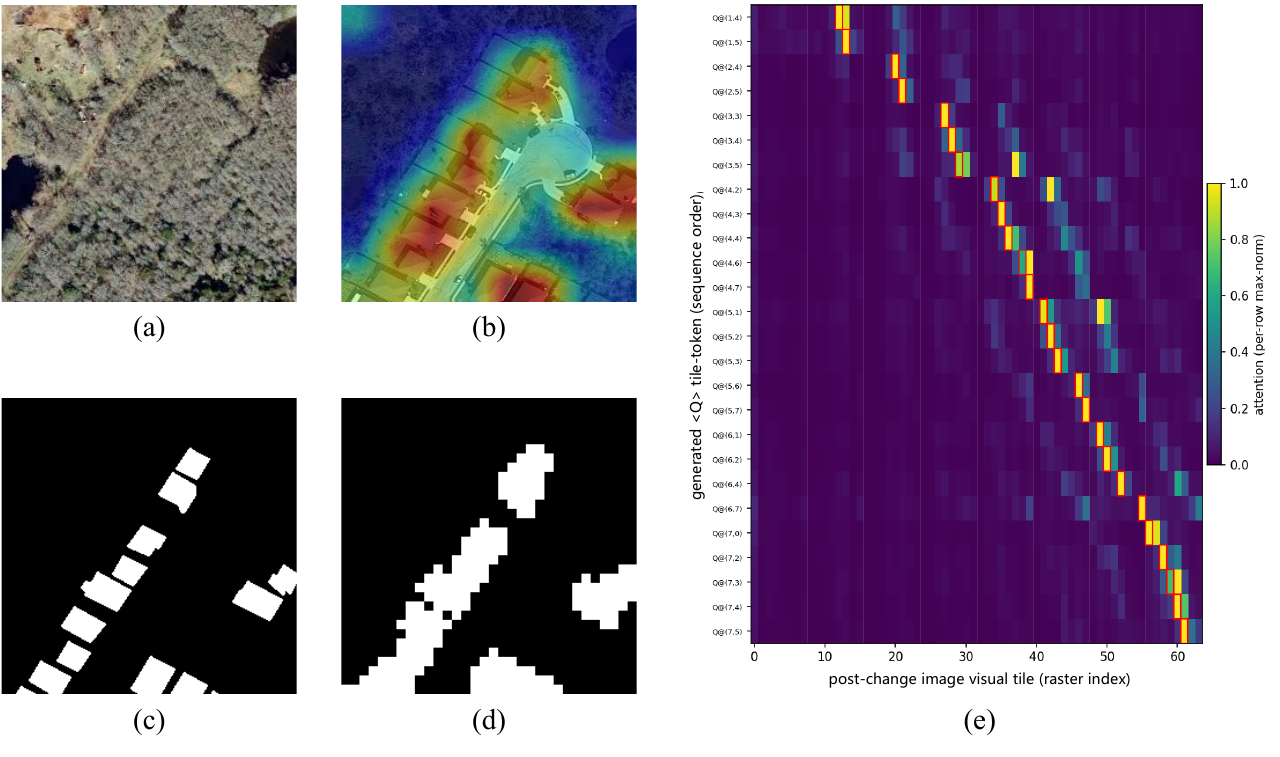}
    \caption{Qualitative attention visualization example. (a) Pre-change image; (b) post-change image with aggregated attention overlay for the complete segmentation sequence; (c) ground-truth change mask; (d) predicted change mask; (e) token-conditioned attention visualization, where the red box indicates the visual tile corresponding to the generated \textit{QUAKE} token. The token-conditioned attention peaks within the corresponding red-boxed tile, while the aggregated attention concentrates on regions of temporal change and is consistent with both the predicted and ground-truth masks.}
    \label{fig:attention_vis_1}
\end{figure*}

\begin{table*}[!t]
\centering
\begin{tabular}{lcc}
\Xhline{1.2pt}
Stage & Grammar Invalid & Structure Mismatch / Fallback \\ \Xhline{1.2pt}
After Curriculum SFT            & $1.26\%$ & $12.64\%$ \\
After RL & $0.13\%$ & $2.17\%$ \\ \Xhline{1.2pt}
\end{tabular}
\caption{Failure-mode rates on the QUAKE-CoT test split. The pad/crop fallback is invoked exactly when a structure mismatch is detected, so the two columns share a trigger.}
\label{tab:appendix_failure}
\end{table*}

\begin{figure*}[!ht]
    \centering
    \includegraphics[width=0.95\linewidth]{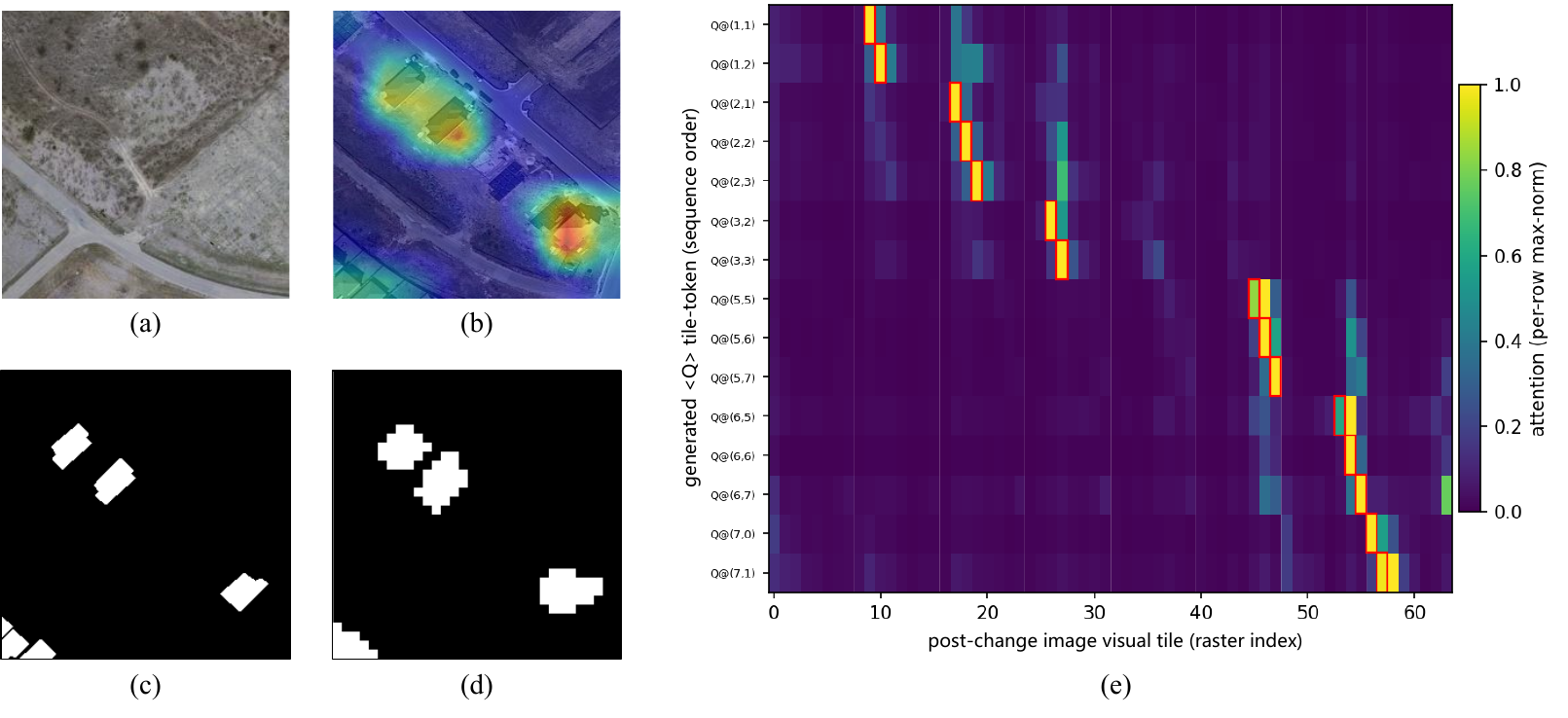}
    \caption{Qualitative attention visualization example. (a) Pre-change image; (b) post-change image with aggregated attention overlay for the complete segmentation sequence; (c) ground-truth change mask; (d) predicted change mask; (e) token-conditioned attention visualization, where the red box indicates the visual tile corresponding to the generated \textit{QUAKE} token. The token-conditioned attention peaks within the corresponding red-boxed tile, while the aggregated attention concentrates on regions of temporal change and is consistent with both the predicted and ground-truth masks.}
    \label{fig:attention_vis_2}
\end{figure*}

\begin{figure*}[!ht]
    \centering
    \includegraphics[width=0.95\linewidth]{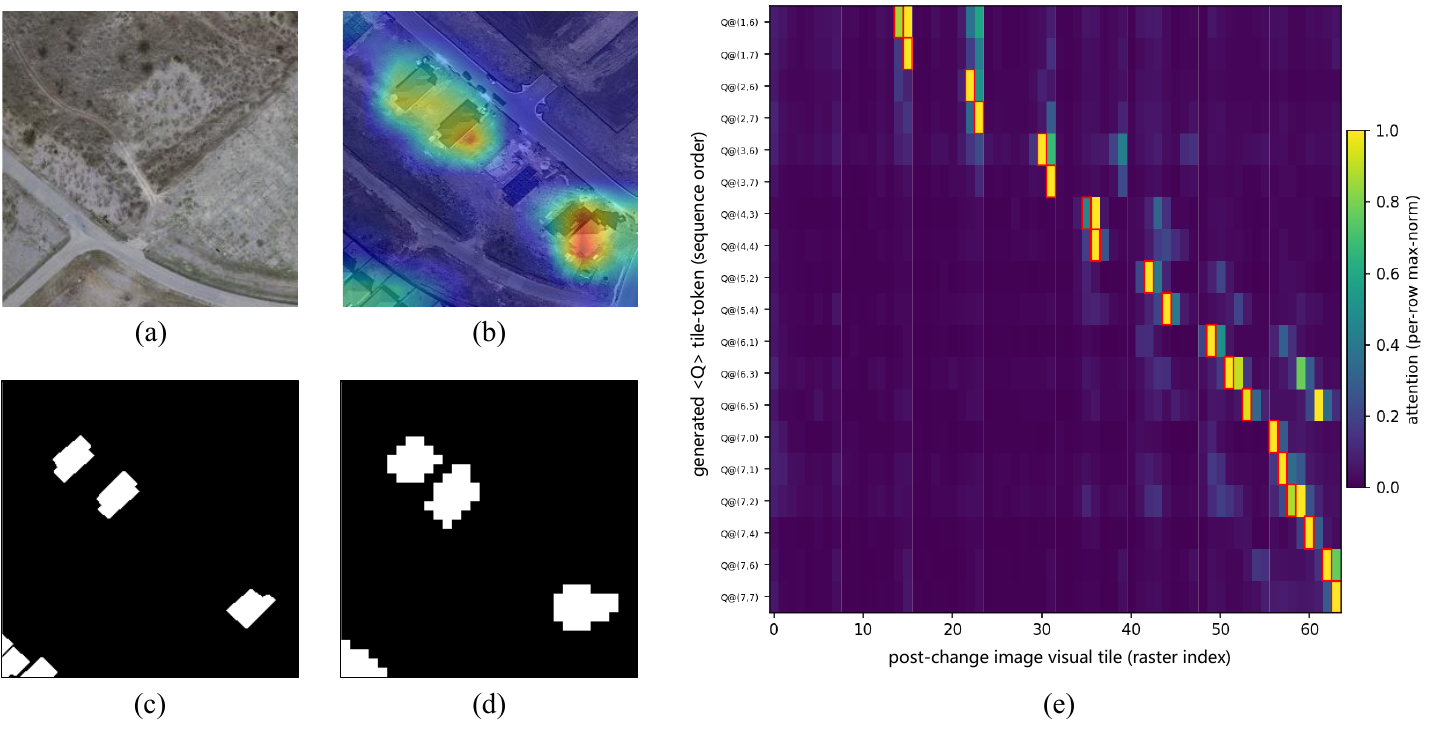}
    \caption{Qualitative attention visualization example. (a) Pre-change image; (b) post-change image with aggregated attention overlay for the complete segmentation sequence; (c) ground-truth change mask; (d) predicted change mask; (e) token-conditioned attention visualization, where the red box indicates the visual tile corresponding to the generated \textit{QUAKE} token. The token-conditioned attention peaks within the corresponding red-boxed tile, while the aggregated attention concentrates on regions of temporal change and is consistent with both the predicted and ground-truth masks.}
    \label{fig:attention_vis_3}
\end{figure*}

\subsection{Qualitative Attention Visualization}
\label{sec:appendix_qualitative}
To assess the spatial grounding of the structured prediction process, we construct two complementary attention views. The \textbf{token-conditioned visualization} projects the attention associated with each generated \textit{QUAKE}
token onto its designated visual tile. The \textbf{aggregated overlay} pools the visual responses over the complete segmentation sequence, providing a holistic view of the evidence supporting the predicted change region.

We present three representative examples in Figure~\ref{fig:attention_vis_1}, Figure~\ref{fig:attention_vis_2}, and Figure~\ref{fig:attention_vis_3}. Across all cases, both attention views concentrate on regions undergoing meaningful temporal changes, while largely unchanged background areas receive weak responses. In the token-conditioned visualization, the red box marks the visual tile corresponding to the generated \textit{QUAKE} token. The peak attention mostly falls within this tile, with substantially lower responses on neighboring and non-corresponding tiles. This self-tile concentration provides direct evidence of the spatial alignment between the \textit{QUAKE} output and the visual representation. The aggregated attention overlay further recovers the overall extent of the changed regions and follows their salient structures and boundaries. Its spatial distribution is consistent with both the predicted masks and the ground-truth annotations, indicating that the model focuses on relevant changes while suppressing irrelevant contextual content. This agreement is maintained across examples with different object scales, shapes, and background configurations, supporting the effectiveness of the \textit{QUAKE}-based representation for grounding structured segmentation predictions in visual change evidence.

\subsection{Parameter Sensitivity Analysis}
\label{sec:appendix_parameter}

We sweep three reward-related hyperparameters around the default configuration used in the main experiments: the Tversky asymmetry $(\alpha,\beta)=(0.3,0.7)$, the reward trade-off $\mu=0.1$ in Eq.~(3), and the KL coefficient $0.04$. As summarized in Tables~\ref{tab:appendix_tversky_sensitivity}--\ref{tab:appendix_kl_sensitivity}, the default setting is consistently the most balanced choice: it yields the strongest recall/F1 trade-off for the Tversky reward, achieves the highest reranker score at $\mu=0.1$, and provides the best overall balance under KL regularization. Although some nearby settings improve precision in isolation, they generally do so at the expense of recall, F1, or reranker performance.

\section{Failure Mode Analysis}
\label{sec:appendix_failure}

We quantify three categories of generation failure that can compromise mask reconstruction. \textbf{Grammar invalidity} occurs when the \texttt{<seg>} span cannot be parsed under $\mathcal{G}_Q$ (e.g., malformed quadtree node, run-length count below $2$, or unmatched delimiters). \textbf{Structure mismatch} occurs when the parsed grid does not match the expected $H/32 \times W/32$ dimensions. Whenever a structure mismatch is detected, the decoder falls back to a \textbf{pad/crop alignment} that pads or truncates the recovered grid to the target shape; the trigger rate of this fallback therefore coincides with the structure-mismatch rate.

Table~\ref{tab:appendix_failure} reports these rates measured on the held-out test split after the two-stage curriculum (S1+S2) and after grammar-gated dual-reward RL (S1+S2+$R_q$+$R_t$). Curriculum supervision already keeps grammar invalidity at $1.26\%$, indicating that the syntax is largely internalized during SFT. Reinforcement learning further reduces grammar invalidity to $0.13\%$, an order-of-magnitude drop attributable to the grammar gate in $R_q$ that assigns zero reward to ungrammatical rollouts. Structure mismatch decreases from $12.64\%$ to $2.17\%$ over the same transition, showing that the structure score in $R_q$ effectively penalizes grid-dimension drift even though the grammar gate alone does not enforce it.

\section{Prompt Templates}
\label{sec:appendix_prompt}

\begin{figure*}[!t]
    \centering
    \includegraphics[width=\linewidth]{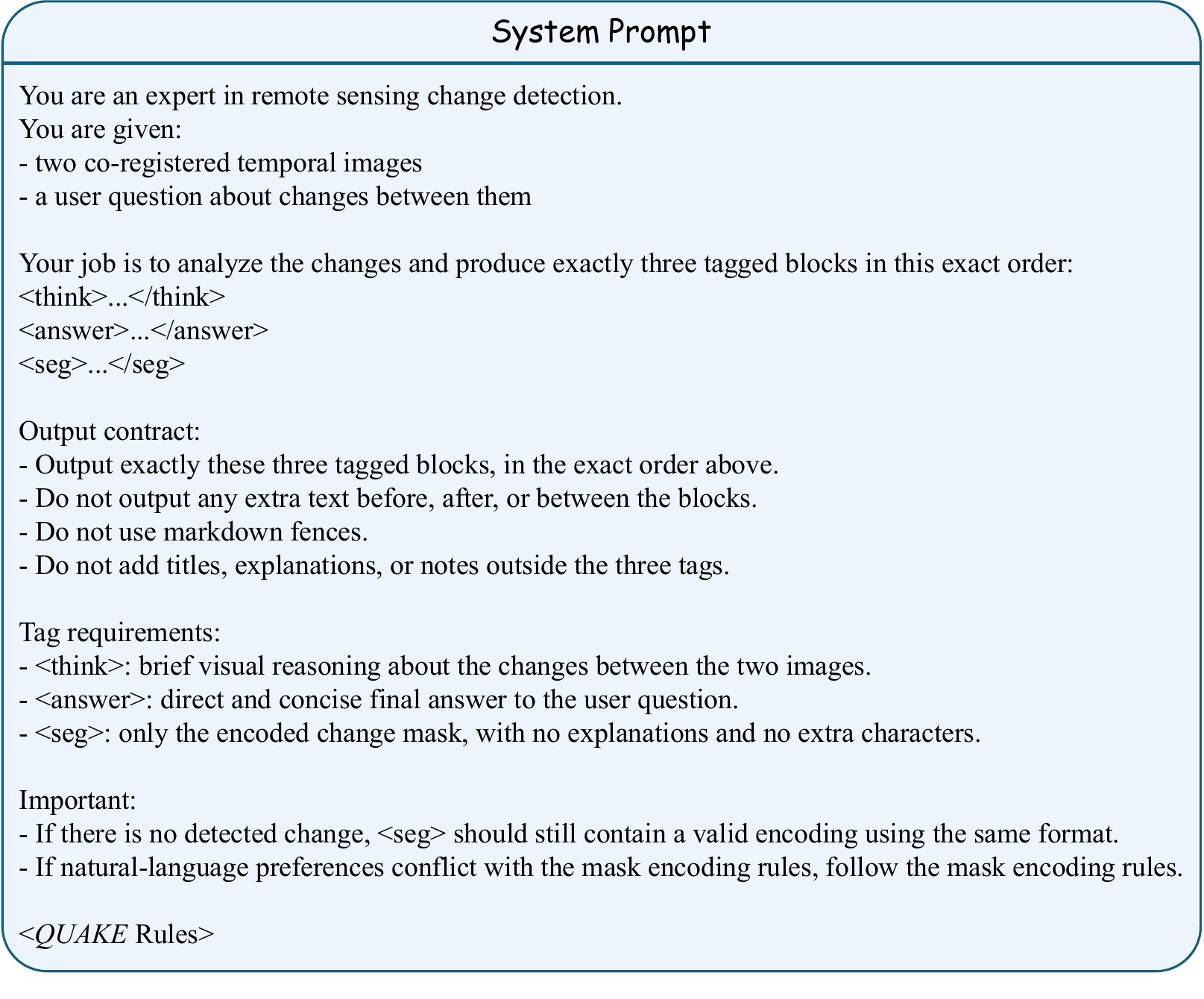}
    \caption{System prompt prepended to all training and inference inputs of \textsc{QUAKE-CD}.}
    \label{fig:sys_prompt_1}
\end{figure*}

\begin{figure*}[!t]
    \centering
    \includegraphics[width=0.9\linewidth]{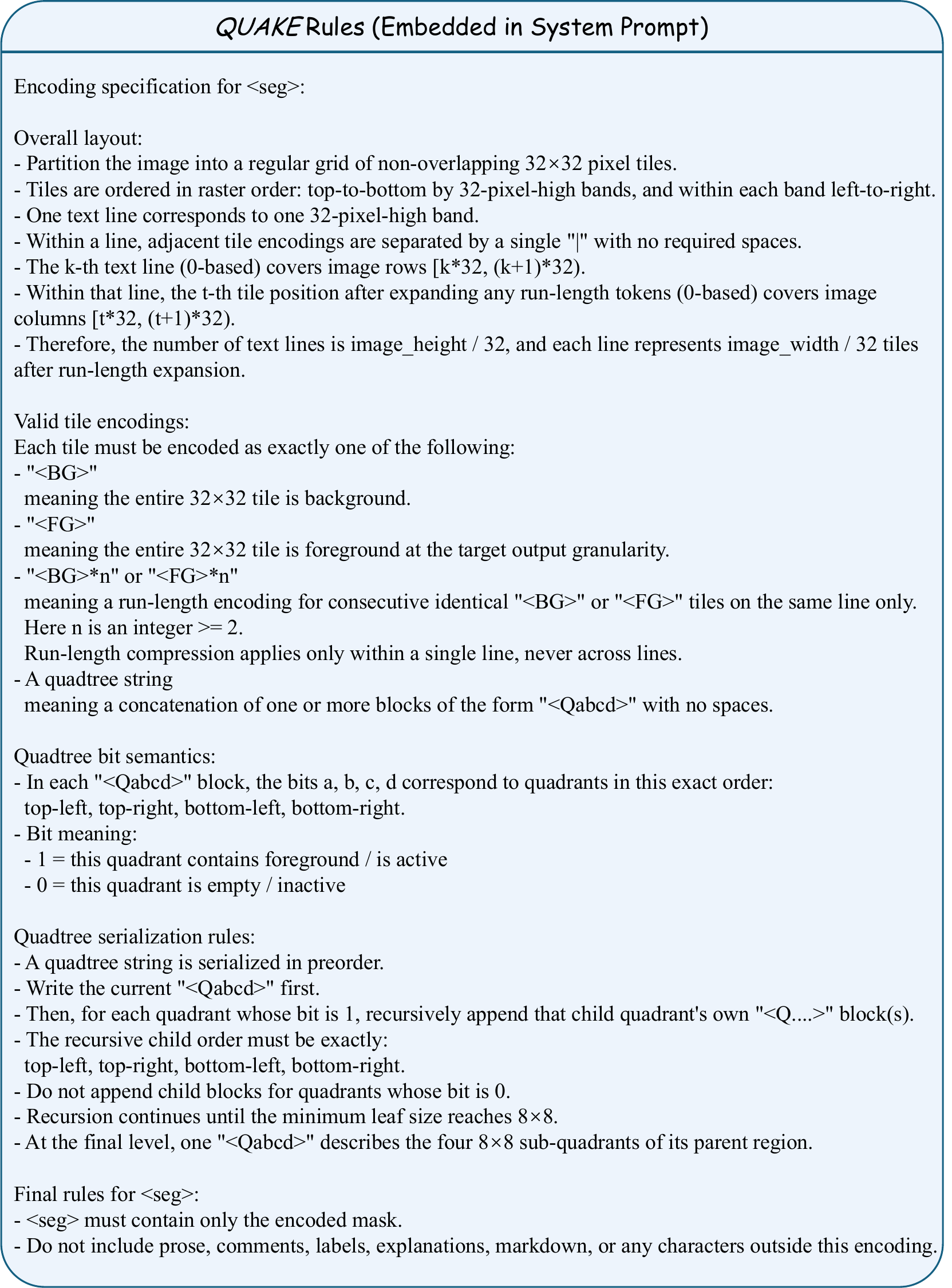}
    \caption{QUAKE grammar rules embedded into the system prompt to constrain the model's output schema.}
    \label{fig:sys_prompt_2}
\end{figure*}

\begin{figure*}[!t]
    \centering
    \includegraphics[width=\linewidth]{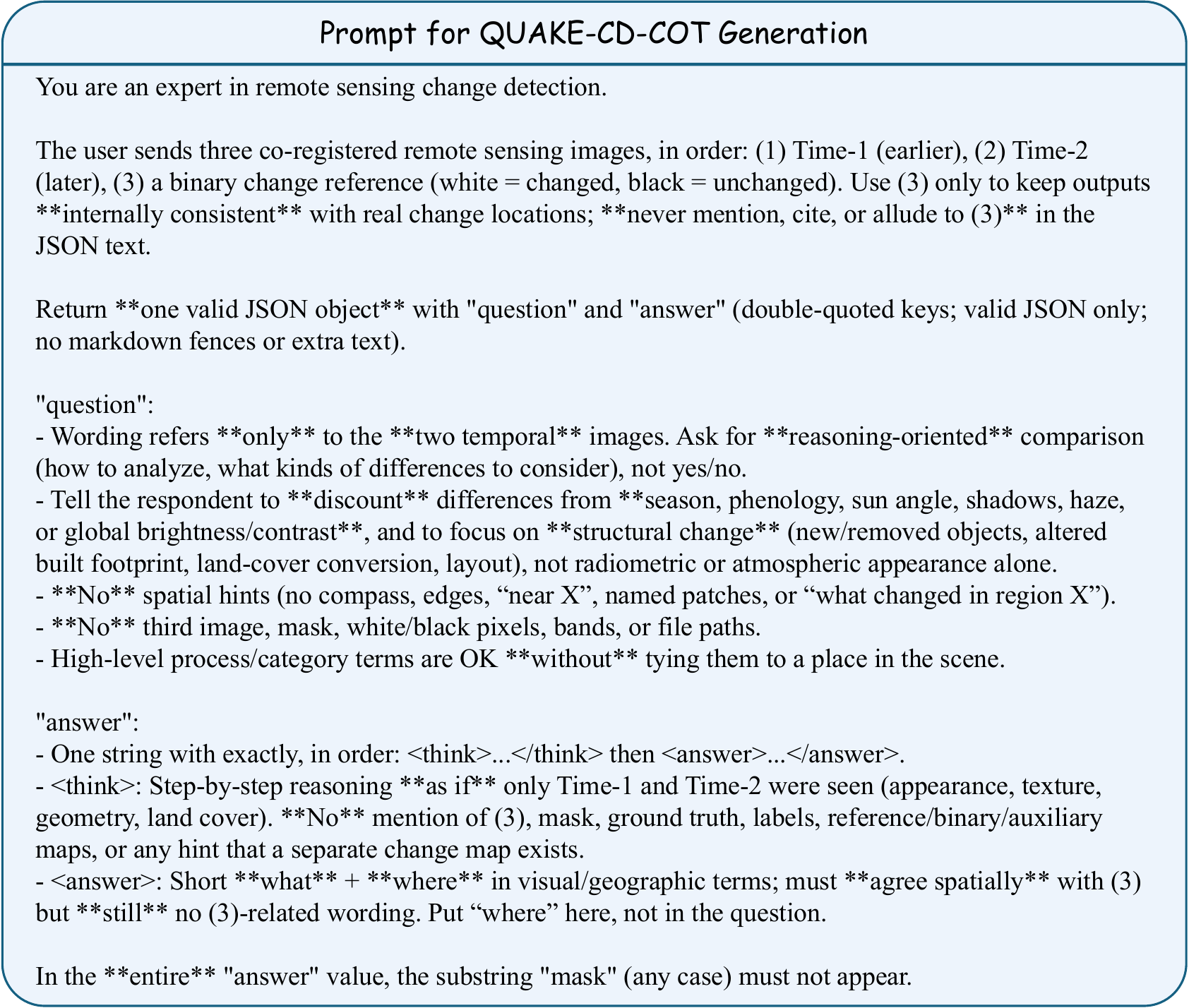}
    \caption{Prompt issued to the teacher model to distill \texttt{<think>}/\texttt{<answer>} traces from a bi-temporal image pair during QUAKE-CoT construction.}
    \label{fig:qa_prompt}
\end{figure*}

\begin{figure*}[!t]
    \centering
    \includegraphics[width=\linewidth]{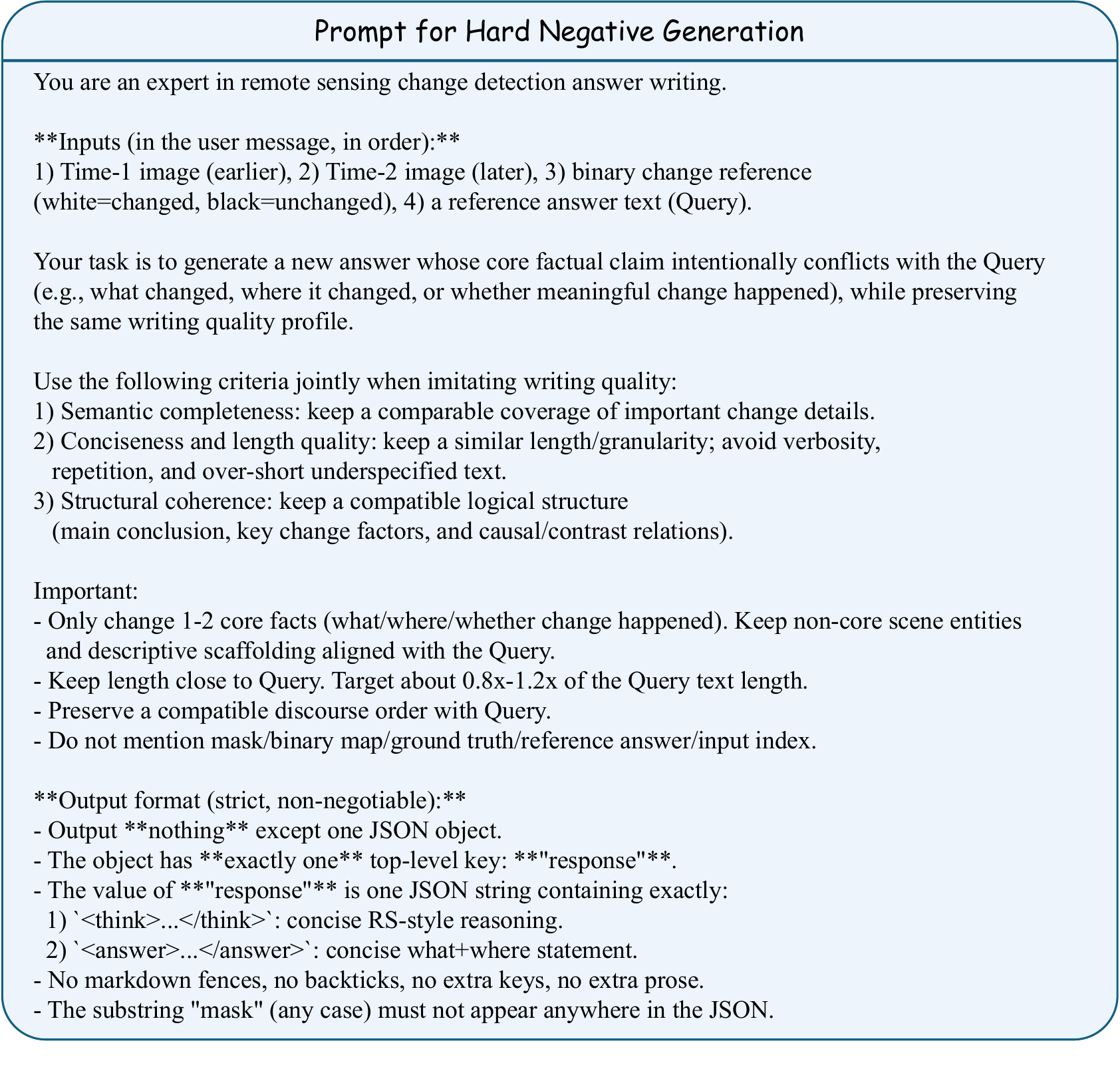}
    \caption{Prompt issued to Gemini-3.1-Pro-Thinking-Preview to synthesize hard negatives for the preference hierarchy in \S\ref{sec:appendix_rm}.}
    \label{fig:hard_negative_prompt}
\end{figure*}

\begin{figure*}[!t]
    \centering
    \includegraphics[width=\linewidth]{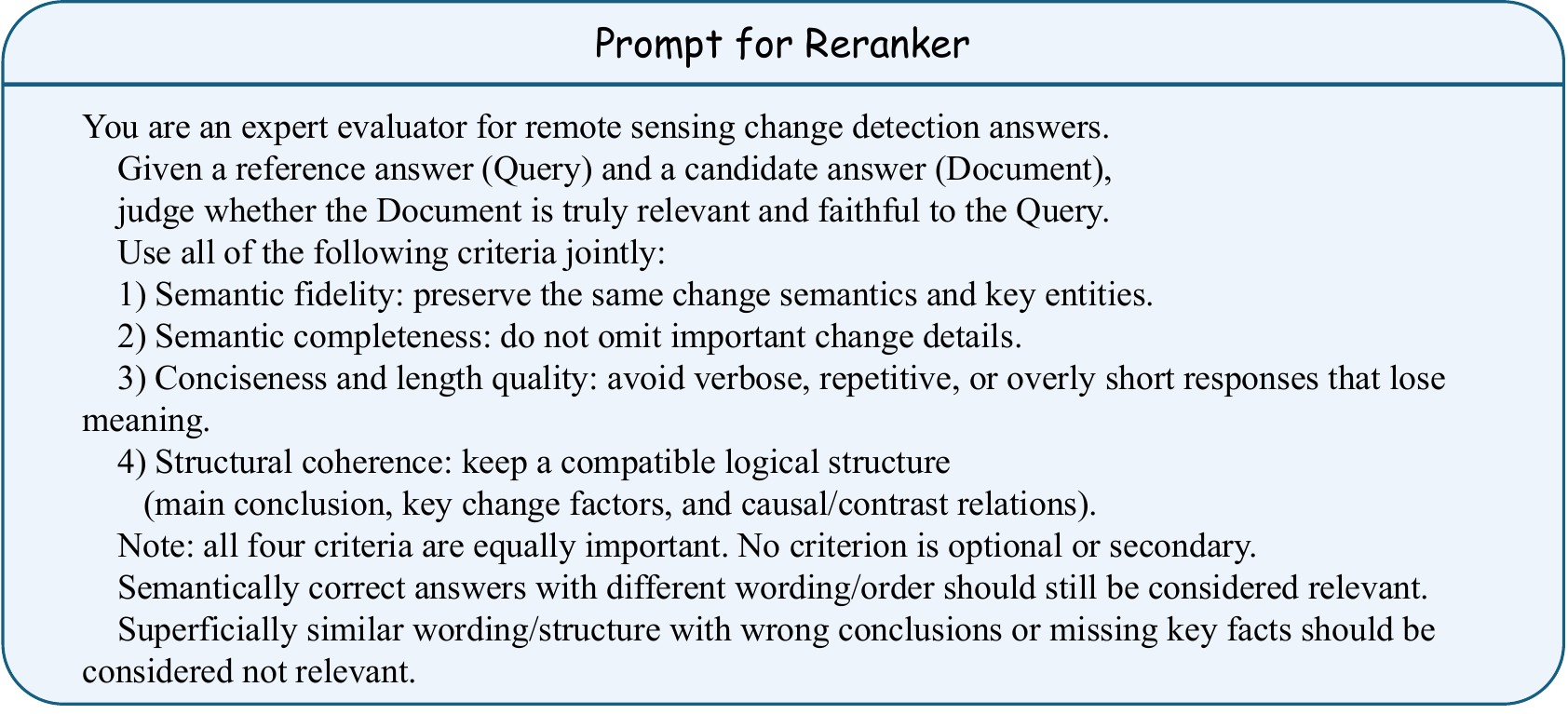}
    \caption{Prompt issued to Qwen3-VL-Reranker to score candidate responses during preference-pair selection.}
    \label{fig:reranker_prompt}
\end{figure*}

\begin{figure*}[!t]
    \centering
    \includegraphics[width=\linewidth]{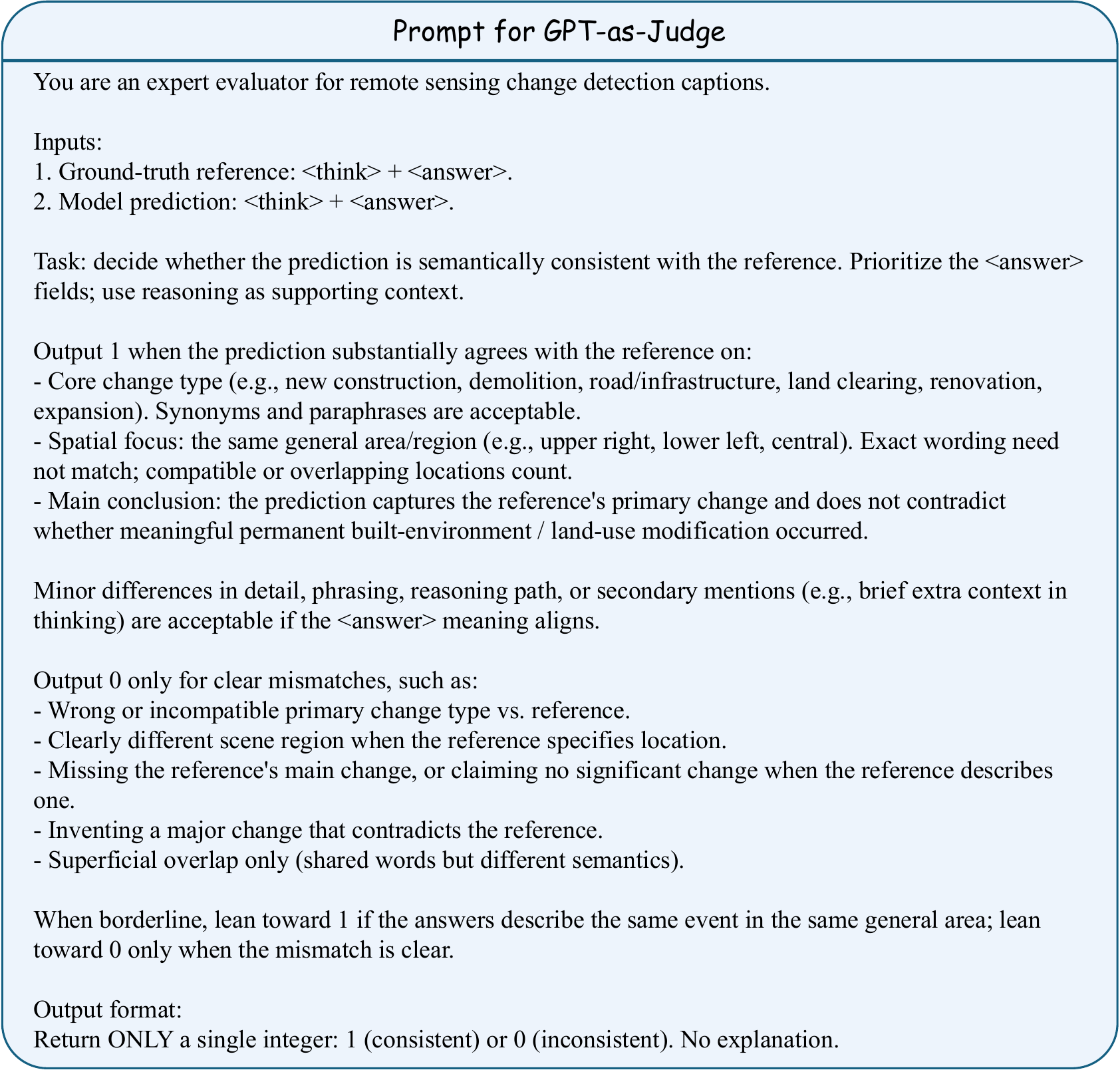}
    \caption{Prompt issued to GPT-5.5 for pairwise quality scoring reported in \S\ref{sec:experiments}.}
    \label{fig:gpt_judge_prompt}
\end{figure*}

Figures~\ref{fig:sys_prompt_1}--\ref{fig:gpt_judge_prompt} reproduce every prompt used in this work. Data construction prompts (Figs.~\ref{fig:sys_prompt_1}, \ref{fig:sys_prompt_2}, \ref{fig:qa_prompt}, \ref{fig:hard_negative_prompt}) cover the system instruction, QA distillation, and hard-negative synthesis described in \S\ref{sec:appendix_dataset} and \S\ref{sec:appendix_rm}. Evaluation prompts (Figs.~\ref{fig:reranker_prompt}, \ref{fig:gpt_judge_prompt}) cover the reranker used during preference-pair selection and the GPT judge used in \S\ref{sec:experiments}.

\end{document}